\documentclass[11pt]{article}
\usepackage{arxiv}
\DeclareUnicodeCharacter{00A7}{\S}
\DeclareUnicodeCharacter{00B0}{\textdegree}
\DeclareUnicodeCharacter{00B1}{\ensuremath{\pm}}
\DeclareUnicodeCharacter{00D7}{\ensuremath{\times}}
\DeclareUnicodeCharacter{0394}{\ensuremath{\Delta}}
\DeclareUnicodeCharacter{03C3}{\ensuremath{\sigma}}
\DeclareUnicodeCharacter{2026}{\ldots}
\DeclareUnicodeCharacter{2192}{\ensuremath{\rightarrow}}
\DeclareUnicodeCharacter{2212}{\ensuremath{-}}
\DeclareUnicodeCharacter{2248}{\ensuremath{\approx}}
\DeclareUnicodeCharacter{2265}{\ensuremath{\ge}}
\usepackage[table]{xcolor}
\definecolor{TietTableStripe}{HTML}{F3F6FA}
\definecolor{TietTableRule}{HTML}{CBD5E1}
\makeatletter
\@ifpackageloaded{caption}{}{\usepackage{caption}}
\AtBeginDocument{%
\ifdefined\contentsname
  \renewcommand*\contentsname{Table of contents}
\else
  \newcommand\contentsname{Table of contents}
\fi
\ifdefined\listfigurename
  \renewcommand*\listfigurename{List of Figures}
\else
  \newcommand\listfigurename{List of Figures}
\fi
\ifdefined\listtablename
  \renewcommand*\listtablename{List of Tables}
\else
  \newcommand\listtablename{List of Tables}
\fi
\ifdefined\figurename
  \renewcommand*\figurename{Figure}
\else
  \newcommand\figurename{Figure}
\fi
\ifdefined\tablename
  \renewcommand*\tablename{Table}
\else
  \newcommand\tablename{Table}
\fi
}
\@ifpackageloaded{float}{}{\usepackage{float}}
\floatstyle{ruled}
\@ifundefined{c@chapter}{\newfloat{codelisting}{h}{lop}}{\newfloat{codelisting}{h}{lop}[chapter]}
\floatname{codelisting}{Listing}

\makeatother
\makeatletter
\@ifpackageloaded{caption}{}{\usepackage{caption}}
\@ifpackageloaded{subcaption}{}{\usepackage{subcaption}}
\makeatother

\title{MDIA: A Multi-Agent Diagnostic Intelligence Pipeline on
HealthBench Professional}

\author{%
Roberto Cruz%
\thanks{\texttt{roberto.cruz@tiet.ai}}%
 \qquad %
David Rey-Blanco%
\thanks{\texttt{david.rey@tiet.ai}}%
\\[0.4em]
\small TietAI%
}

\date{2026-05-15}

\begin{document}
\maketitle

\begin{abstract}
\noindent Most reported gains on agentic-LLM clinical benchmarks are
often attributed to prompt engineering, yet our results suggest that
larger improvements can come from architectural and engine-level design.
We present MDIA, a Multi-agent Diagnostic Intelligence Agent implemented
as a 7-node specialty-routed clinical reasoning graph, on the full
HealthBench Professional benchmark (n = 525), on a non-fine-tuned LLM.
MDIA achieves 0.6272 under OpenAI's GPT-5.4-2026-03-05, which is +3.72
pp above the performance of OpenAI's ChatGPT for Clinicians. The
experimental work shows that performance lift is attributable to system
architecture: specialty routing, multi-turn context preservation,
drug-state safety gating, site-filtered search, length-aware synthesis,
and engine-level reliability. These findings support the view that
agentic clinical benchmark performance is shaped both by the underlying
foundation model and the orchestration architecture. Nevertheless, we
also noticed notable differences when using other models as a grader; in
particular, when using Gemini 2.5 Pro, MDIA scored 0.6585, which
suggests that the choice of grader is a source of variability. Robust
evaluation of LLMs would therefore require assessment across several
independent grader models.
\end{abstract}

\bigskip

\section{Introduction}\label{introduction}

Large language models (LLMs) are entering clinical practice at a pace
that makes systematic evaluation essential \citep{lee2023nejmgpt4}. The
release of OpenAI's HealthBench Professional \citep{healthbenchpro2026}
--- 525 rubric-graded cases drawn from 15,079 real clinician
conversations --- provides the field's most rigorous open benchmark for
this task. Its design is deliberately adversarial: cases are stratified
across 21 specialties, include both good-faith and red-teaming
scenarios, and a significant fraction (22 \%) require a model to
integrate follow-up turns rather than answer a single isolated question.
Top-line scores from the original paper establish demanding reference
points: physician-written responses at 0.437, GPT-5.4 single-agent at
0.481, and OpenAI's best system, ChatGPT for Clinicians, at 0.590.

The literature on medical AI agents identifies a consistent pattern:
agent architectures with tool use substantially outperform single-prompt
baselines. A systematic review of 20 clinical agent studies found that
all agent architectures outperformed their corresponding baseline LLMs,
with a median improvement of +53 pp for single-agent tool-calling
systems \citep{aiagentssysrev2025}. Multi-agent frameworks further
extend this advantage by assigning specialised roles --- a paradigm
validated by MedAgents \citep{medagents2024}, which demonstrated
zero-shot state-of-the-art on MedQA through specialised collaborative
discussion. Despite these results, multi-agent clinical pipelines have
not yet been evaluated on HealthBench Professional, and no prior work
has examined how the benchmark's multi-turn conversation structure
interacts with common evaluation harness design choices.

This paper addresses both gaps with \textbf{MDIA} (Multi-agent
Diagnostic Intelligence Agent), a coordinated 7 agent specialty-routed
Directed Acyclic Graph (DAG) with shared memory: (1) an intake
orchestrator with 14 medical tools (PubMed, DailyMed, UMLS, ICD-10,
drug-state safety gate, site-filtered web search, and more) collects a
structured clinical dossier; (2) a specialty classifier branches the
work to one of three domain-expert reasoners ( (3) gastroenterology, (4)
ophthalmology, (5) neurology) or a (6) generalist path; an (7) output
synthesizer produces the final response; and a verifier performs a final
safety and format check.

\begin{figure}

\centering{

\pandocbounded{\includegraphics[keepaspectratio]{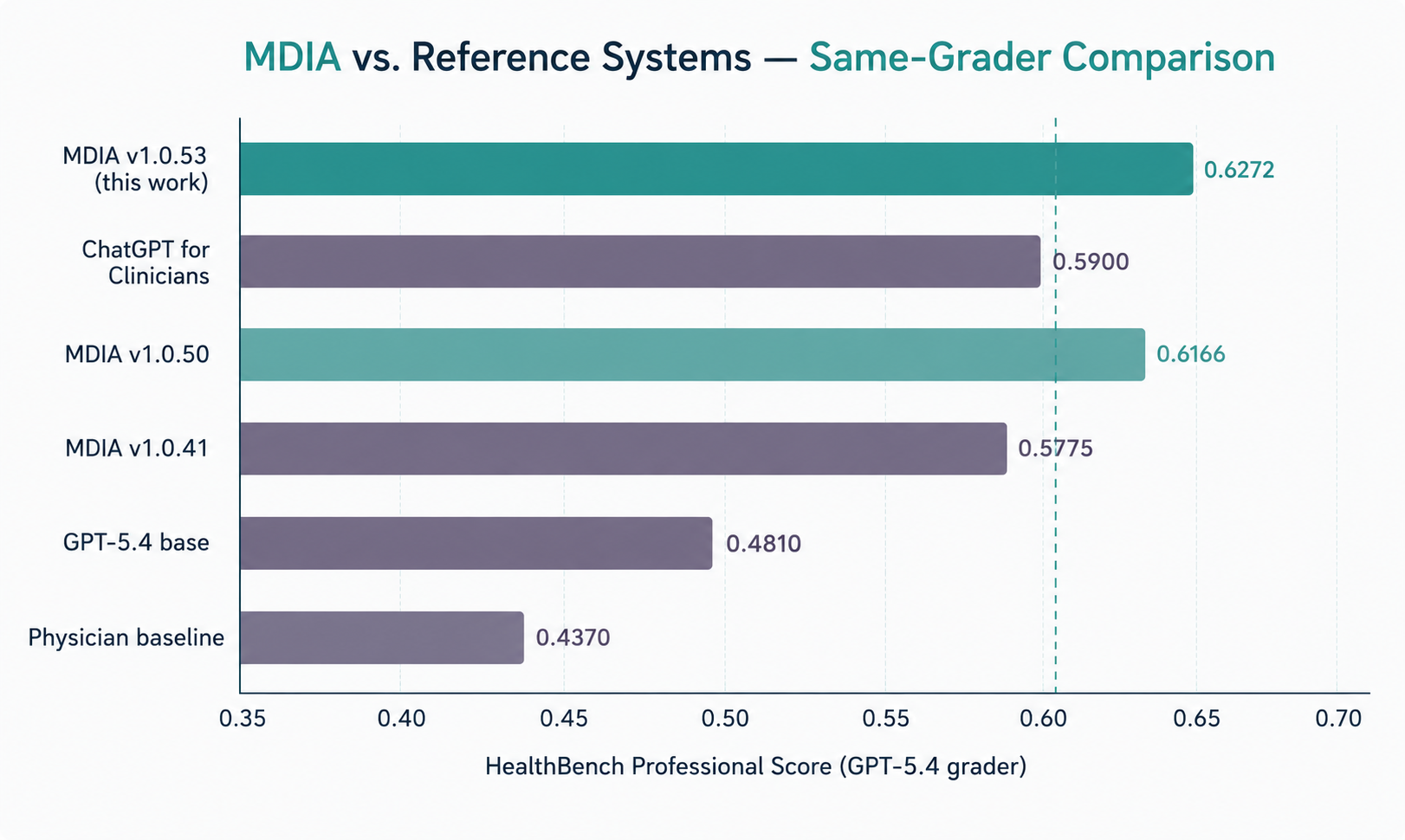}}

}

\caption{\label{fig-overall-comparison}Overall comparison of MDIA
versions against OpenAI reference systems under GPT-5.4 grading. MDIA
v1.0.53 (0.6272) surpasses ChatGPT for Clinicians (0.590) by 3.72
pp.~Source: internal}

\end{figure}

The headline result shows our agent achieves a total score of
\textbf{0.6272} on GPT-5.4-2026-03-05 low reasoning \citep{gpt54} ---
OpenAI's own grader --- on all 525 samples. The same-instrument
comparison uses the GPT-5.4 score: \textbf{+14.62 pp over the GPT-5.4
single-agent baseline} (0.6272 vs 0.481), \textbf{+19.02 pp over
physician-written responses} (0.6272 vs 0.437), and nominally
\textbf{+3.72 pp ahead of ChatGPT for Clinicians} (0.6272 vs 0.590). The
last margin lies within bootstrap σ (≈ 0.023) and should be treated as
directional rather than decisive --- OpenAI does not publish confidence
intervals for ChatGPT for Clinicians, precluding a significance test.
Additionally, if OpenAI's system supports multi-turn context (the
flatten strategy used in their evaluation is undocumented), the
effective gap may differ from the nominal 3.72 pp.~All lift comes from
architectural design on the open TietAI Hydra Platform
\citep{hydraplatform2026}, not from proprietary data or model access.
The overall same-grader comparison is summarized in
Figure~\ref{fig-overall-comparison}.

A secondary finding with implications beyond this work:
\textbf{conversation flatten strategy moves the HealthBench Pro headline
by approximately 6 pp at n=525}. The standard simple-evals harness
passes only the last user message to the agent. Of HealthBench Pro's 525
cases, 115 (22 \%) contain follow-up user turns whose context is
silently discarded by this default. Switching to a full multi-turn pass
lifts the score from 0.6102 to 0.6598 Pro-graded --- with zero changes
to the agent --- because the rubric grades the agent's ability to
address the complete conversation, not just the final question. This
effect is quantified in Section~\ref{sec-methodology} and
Section~\ref{sec-multiturn-finding}. An important caveat for the ChatGPT
for Clinicians comparison: OpenAI has not documented the flatten
strategy used for that system's evaluation. If it too uses multi-turn
context, their published 0.590 score already captures this advantage,
and the nominal +3.72 pp gap reflects other architectural differences.

To further assess the robustness of our evaluation, we also graded the
results using an alternative LLM judge, Gemini 2.5 Pro with MDIA v1.0.50
attaining a rubric of 0.6771 ± 0.0204. While the overall score
magnitudes and relative ranking remained broadly consistent, we observed
differences in how individual responses were scored. These discrepancies
highlight the limitations of relying on a single LLM as a judge and
suggest that multi-grader evaluation may provide a more reliable
assessment. This conclusion is consistent with prior work on
LLM-as-judge limitations and multi-agent evaluation, including
MT-Bench/Chatbot Arena and ChatEval, which respectively document judge
bias and propose multi-agent referee teams to improve alignment with
human assessment \citep{zheng2023judging, chan2024chateval}.

Our paper makes four contributions to clinical agent evaluation and
deployment: (1) a working multi-agent clinical pipeline that exceeds
OpenAI's flagship system under their own grader; (2) a multi-turn
evaluation finding that all future HealthBench Pro reporters should
account for; (3) five engine-level reliability fixes in the Hydra
Platform subagent-graph executor that recovered \textasciitilde3-4 pp
previously lost to infrastructure flakiness; and (4) the first
end-to-end validation of the 7-node graph architecture via the correct
graph endpoint.

This paper is organized as follows: firstly, we describe the overall
architecture of the agent; second, we outline the methodological paths
explored during development; third, we evaluate the configurations from
v1.0.27 to v1.0.53; and finally summarizes the lessons learned
throughout the model construction process.

\section{Background and motivation}\label{background-and-motivation}

Early medical LLM benchmarks mainly used examination-style datasets to
assess biomedical knowledge and structured reasoning, including MedQA
\citep{jin2021medqa}, MedMCQA \citep{pal2022medmcqa}, PubMedQA
\citep{jin2019pubmedqa}, and medical subsets of MMLU
\citep{hendrycks2020measuring}. These benchmarks enabled reproducible
comparison and helped show that frontier LLMs could approach or exceed
physician passing thresholds, as in the Med-PaLM family
\citep{singhal2023medpalm, singhal2025medpalm2}. However, they primarily
test static question-answering rather than clinical deployment
capabilities such as safety, communication, uncertainty management, or
longitudinal decision-making, motivating newer workflow-oriented
evaluations.

OpenAI's HealthBench \citep{healthbench2025} moved evaluation toward
realistic healthcare interactions using 5,000 multi-turn clinical and
patient-facing conversations graded with physician-authored rubrics.
HealthBench Professional \citep{healthbenchpro2026} extends this
approach to clinician workflows, curating 525 tasks from
physician-generated conversations across 50 countries and 26
specialties, with emphasis on care consultation, documentation, medical
research, specialist adjudication, and adversarial cases. Despite these
advances, such benchmarks remain technical proxies rather than clinical
validation tools, and issuer-related conflicts may arise when the
benchmark provider is also a model vendor. Therefore, benchmark results
should be interpreted as standardized technical evidence, not as
substitutes for prospective clinician-led trials, regulatory-grade
evaluation, or real-world outcome studies.

HealthBench professional is a rubric-graded benchmark drawn from real
clinician chats --- 15,079 initial conversations distilled to 525
high-signal cases via stratified sampling. Each case carries
multi-criterion rubrics with weighted positive and negative items, and
the scoring formula is \(sum(earned\_points) / sum(positive\_points)\)
per example, averaged across the 525 samples and clipped to {[}0, 1{]}.

The reference numbers from OpenAI's paper are summarized in
Table~\ref{tbl-openai-reference-baselines}.

\begin{longtable}[]{@{}lrl@{}}
\caption{OpenAI HealthBench Professional reference
baselines.}\label{tbl-openai-reference-baselines}\tabularnewline
\toprule\noalign{}
System & Score & Coverage \\
\midrule\noalign{}
\endfirsthead
\toprule\noalign{}
System & Score & Coverage \\
\midrule\noalign{}
\endhead
\bottomrule\noalign{}
\endlastfoot
Physician-written baseline & 0.437 & n = 525 \\
GPT-5.4 base (single-agent) & 0.481 & n = 525 \\
ChatGPT for Clinicians (best published) & 0.590 & n = 525 \\
\end{longtable}

Our goal is to build a multi-agent system on a general purpose LLM that
meets or exceeds the OpenAI flagship under their own grader, without
fine-tuning, on a small team (one engineer + a 14-tool platform), and
with full reproducibility --- graph definition, prompts, and per-sample
grader transcripts publishable. Recent benchmarks (AgentClinic
\citep{agentclinic2026}, MedAgentBench \citep{medagentbench2025},
PhysicianBench \citep{physicianbench2026}) have established that even
frontier models struggle in the sequential, tool-using clinical settings
that HealthBench Pro approximates; a systematic review of 20 agent
studies found that all agent architectures outperformed their baseline
LLMs, with a median improvement of +53 pp for single-agent tool-calling
systems \citep{aiagentssysrev2025}.

A non-obvious requirement we discovered during the work was the
importance of respecting the conversation structure of the dataset.
HealthBench Pro's \emph{conversation.messages} field contains follow-up
turns; 115 / 525 cases (22 \%) include a second user turn that refines
the question (e.g.~``and now show me a table of permitted foods''). The
simple-evals reference loop flattens to the last user turn only ---
silently dropping the context the rubric grades against. We discuss this
in Section~\ref{sec-methodology} and quantify the impact in
Section~\ref{sec-multiturn-finding}.

\section{Architecture}\label{architecture}

\begin{figure}

\centering{

\pandocbounded{\includegraphics[keepaspectratio]{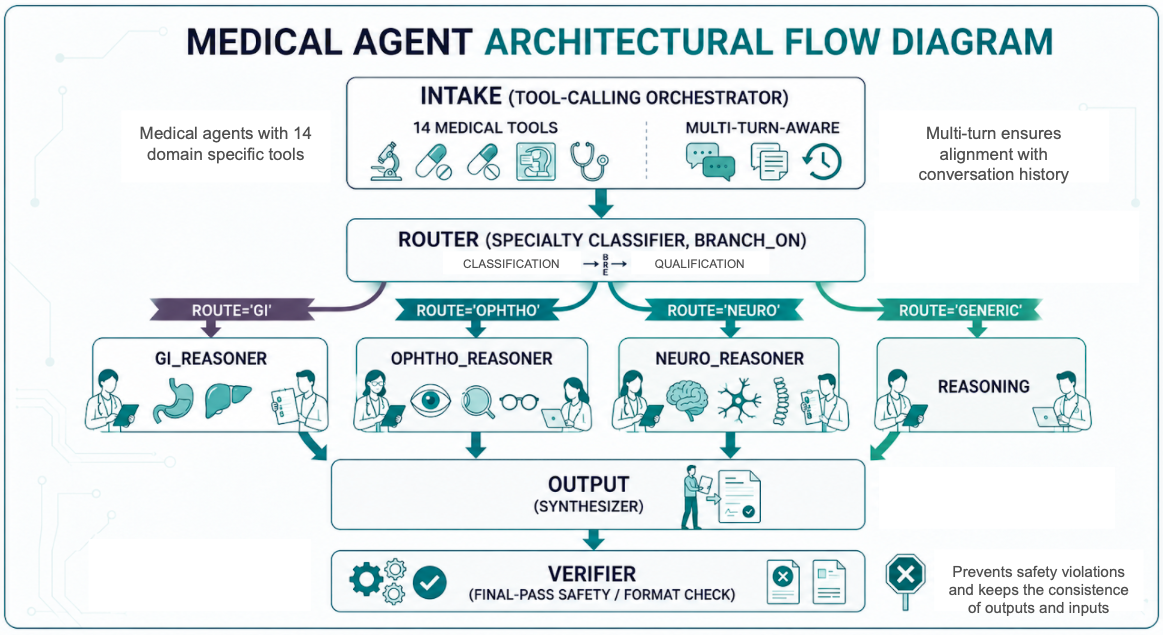}}

}

\caption{\label{fig-architecture}MDIA Architecture}

\end{figure}%

MDIA is implemented as a 7-node, specialty-routed directed acyclic graph
executed by the agentic module of TietAI's Hydra Hydra Graph Engine
\footnote{Hydra Graph Engine is one of the agent runtimes within Hydra
  Platform developed by the company Tiet AI} on Google Gemini family:
2.5 Pro \citep{gemini25pro} and 3.1 Pro \citep{gemini31pro}. The graph
structure shown in Figure~\ref{fig-architecture} displays the nodes and
responsibilities listed in Table~\ref{tbl-mdia-graph-nodes}. Execution
begins with tool-set invocation, task-type classification, and ancillary
data retrieval, after which the input and retrieved context are routed
to the most appropriate expert agent. The resulting outputs are then
collected, curated, and validated for safety and formatting.

\renewcommand{\arraystretch}{1.25}
\begingroup
\small

\begin{longtable}[]{@{}
  >{\raggedright\arraybackslash}p{(\linewidth - 2\tabcolsep) * \real{0.2000}}
  >{\raggedright\arraybackslash}p{(\linewidth - 2\tabcolsep) * \real{0.8000}}@{}}
\caption{MDIA graph nodes and
responsibilities.}\label{tbl-mdia-graph-nodes}\tabularnewline
\toprule\noalign{}
\begin{minipage}[b]{\linewidth}\raggedright
Node
\end{minipage} & \begin{minipage}[b]{\linewidth}\raggedright
Role
\end{minipage} \\
\midrule\noalign{}
\endfirsthead
\toprule\noalign{}
\begin{minipage}[b]{\linewidth}\raggedright
Node
\end{minipage} & \begin{minipage}[b]{\linewidth}\raggedright
Role
\end{minipage} \\
\midrule\noalign{}
\endhead
\bottomrule\noalign{}
\endlastfoot
\textbf{Intake} & Tool-calling research orchestrator. Calls 14 medical
tools (PubMed, Europe PMC, ClinicalTrials, DailyMed, CIMA, UMLS, ICD-10,
drug-state safety check, medical calculator, web search,\ldots) and
emits a structured dossier. Multi-turn-aware: if
\texttt{conversation.messages} has follow-up user turns, those carry
through to the dossier. \\
\textbf{Router} & Specialty classifier. Reads the dossier, emits
\texttt{\{"route":\ "...",\ "route\_reason":\ "..."\}}. \textbar{} \\
\textbf{Gi\_reasoner,} \textbf{Ophtho\_reasoner,}
\textbf{Neuro\_reasoner} & Specialty-tuned clinical reasoners with
curated anchor knowledge (Glasgow-Blatchford, Forrest, MELD-Na, Tokyo
Guidelines, NIHSS, tPA-window, Hunt-Hess, House-Brackmann, etc.). \\
\textbf{Reasoning} & Generalist reasoner (covers cards, ID, peds,
surgery, etc. --- \textbar{} the long tail). \\
\textbf{Output} & Synthesizer. Turns the reasoner's brief into a
user-facing response with a length target (2000--3000 chars typical,
4000 hard cap). \\
\textbf{Verifier} & Final-pass safety / format check. \textbar{} \\
\end{longtable}

\endgroup

The graph is published as an immutable, semver-pinned definition
(versions v1.0.27, v1.0.36, v1.0.40, \ldots, v1.0.50) with engine-level
reproducibility guarantees. Specific model assignments per node are
configuration, versioned with the graph; this paper's results were
obtained with the model fleet described in
Section~\ref{sec-reasoner-upgrade}.

\subsection{Why specialty routing (and why three branches, not twenty
eight)}\label{why-specialty-routing-and-why-three-branches-not-twenty-eight}

In the first iterations, we observed unbalanced scores across
specialties (28 in total), and that more curated specialty-specific
prompts---hereafter referred to as specialty subagents---improved
performance. The initial assumption was therefore that assigning one
agent per specialty would yield the best results. This, however, proved
not to be the case, mainly due to two factors:

\begin{enumerate}
\def\labelenumi{\arabic{enumi}.}
\tightlist
\item
  \textbf{The gap between the best- and worst-performing specialties was
  50 pp}: nephrology scored 0.743, whereas ophthalmology scored 0.243 in
  the v1.0.10 baseline. A generalist reasoner systematically
  underperforms on under-represented specialties
  \citep{lievin2024medqa}.
\item
  \textbf{Ninety-four percent of failed positive-criteria points were
  specialty-knowledge anchors}, rather than refusal-policy issues.
\end{enumerate}

Branching by detected specialty allows a Pro-tier reasoner to load
specialty-specific anchor knowledge that a single global prompt cannot
accommodate. We selected \textbf{GI, Ophthalmology, and Neurology} as
the first three branches based on a per-specialty headroom analysis:
together, these specialties account for 9.5 pp of headroom in the full
benchmark (n = 525) and contain the largest number of previously
identified anchor patterns. Adding cardiology or pediatrics branches
produced diminishing returns at this stage. MedAgents
\citep{medagents2024} demonstrated the value of multidisciplinary LLM
collaboration in medical reasoning, achieving state-of-the-art
performance on MedQA in the zero-shot setting through specialized
role-playing agents; we adopt the same intuition through hardwired
specialty branches rather than dynamic role assignment, an architectural
pattern that mirrors Mixture-of-Agents \citep{wang2024moa}.

\section{Methodology}\label{sec-methodology}

The methodology combines several strategy classes introduced across
successive MDIA versions (shown in Figure~\ref{fig-architecture}).
First, the evaluation harness was corrected to preserve multi-turn
conversations rather than flattening each case to the last user message
(v1.0.40). Second, the clinical graph was hardened with scoped
drug-state safety checks, specialty routing, and a reasoner upgrade
whose value depended on preserving conversation context
(v1.0.27--v1.0.40). Third, the Hydra execution engine was made more
reliable through JSON-fence stripping, retry-on-empty behavior, fallback
messages, per-model location overrides, and thinking-content capture.
Finally, later versions focused on evidence and response-shaping
strategies: search hygiene and citation formatting (v1.0.42--v1.0.46),
full graph-endpoint validation (v1.0.50), and length-aware synthesis
plus verification (v1.0.53). The subsections below separate these
strategies so that benchmark gains can be attributed to harness
behavior, clinical-agent design, engine reliability, search/evidence
quality, and final-answer compression.

\begin{figure}

\centering{

\pandocbounded{\includegraphics[keepaspectratio]{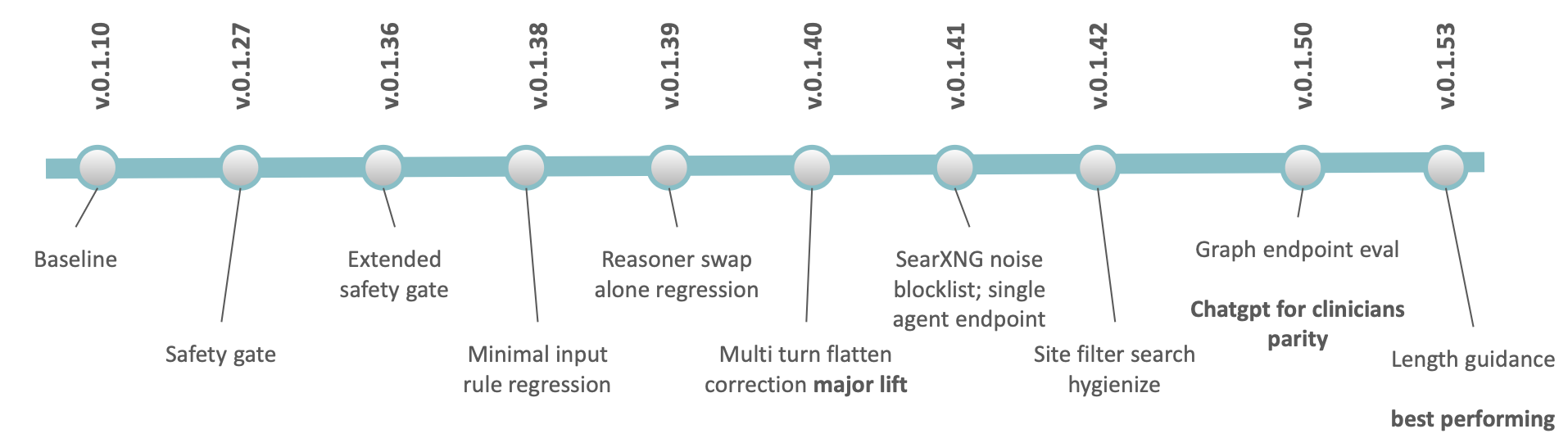}}

}

\caption{\label{fig-methodology-released}MDIA evolution across
experimental releases, showing the main interventions}

\end{figure}%

\subsection{Multi-turn conversation
handling}\label{sec-multiturn-conversation-handling}

HealthBench Pro's dataset stores each example as
\texttt{conversation.messages:\ {[}\{role,\ content\},\ ...{]}}.
\textbf{115 / 525 examples (22 \%)} contain ≥ 2 user turns where the
second turn refines or extends the first (e.g.~``what is the FODMAP
diet'' → ``now give me a table of permitted foods'').

The simple-evals reference harness flattens conversations to the
\textbf{last user turn only} --- sufficient for single-shot QA but
\textbf{silently drops follow-up context that the rubric grades
against}. On a multi-turn case, the agent receives only the second
turn's text (``give me a table''), with no signal that the first turn
defined the topic. We implemented the four flatten strategies detailed
in Table~\ref{tbl-flatten-strategies}:

\begin{longtable}[]{@{}
  >{\raggedright\arraybackslash}p{(\linewidth - 2\tabcolsep) * \real{0.1800}}
  >{\raggedright\arraybackslash}p{(\linewidth - 2\tabcolsep) * \real{0.8200}}@{}}
\caption{Conversation flattening strategies implemented in the
evaluation harness.}\label{tbl-flatten-strategies}\tabularnewline
\toprule\noalign{}
\begin{minipage}[b]{\linewidth}\raggedright
Strategy
\end{minipage} & \begin{minipage}[b]{\linewidth}\raggedright
Behaviour
\end{minipage} \\
\midrule\noalign{}
\endfirsthead
\toprule\noalign{}
\begin{minipage}[b]{\linewidth}\raggedright
Strategy
\end{minipage} & \begin{minipage}[b]{\linewidth}\raggedright
Behaviour
\end{minipage} \\
\midrule\noalign{}
\endhead
\bottomrule\noalign{}
\endlastfoot
last\_user & Reference simple-evals: pass only the last user content. \\
role\_tagged & \texttt{User:\ ...\ /\ Assistant:\ ...\ /\ User:\ ...}
plain prefix. \\
xml &
\texttt{\textless{}turn\ role="user"\textgreater{}...\textless{}/turn\textgreater{}}
block per turn. \\
multiturn & \textbf{Pass the full message list to the agent's invoke
endpoint.} Agent receives \texttt{{[}\{user,\ assistant,\ user\}{]}} and
resolves the follow-up against its own prior reply. \\
\end{longtable}

The new default is \texttt{multiturn}. On single-turn cases (78 \%), it
is byte-equivalent to \texttt{last\_user}. On multi-turn cases, it gives
the agent the full context. \textbf{At n=525, switching from
\texttt{last\_user} to \texttt{multiturn} lifts the Pro-graded score
from 0.6102 → 0.6598 (+5.0 pp) and the GPT-5-graded score from 0.5220 →
0.585 (+6.3 pp).} Single-turn-only subset is essentially flat (within
bootstrap σ); the entire lift comes from the 22 \% multi-turn slice. See
Section~\ref{sec-multiturn-finding}.

This is not an architectural change to MDIA, but it is a fix to the eval
harness that surfaces the agent's existing multi-turn capability. We
argue all HealthBench-Pro reporting should disclose the flatten
strategy, and that \texttt{last\_user} understates any agent that
supports multi-turn. MedMT-Bench \citep{medmtbench2026} independently
confirms that multi-turn medical dialogue is an unsolved problem --- all
17 frontier models score below 60 \% on a 400-case benchmark with an
average of 22 conversation rounds --- underscoring the importance of
correctly preserving conversation context in medical evaluations.

\subsection{Drug-state safety gate}\label{drug-state-safety-gate}

Each reasoner's writing-task branch runs a mandatory \textbf{drug ×
patient-state} check before producing a content brief. This gate played
two roles:

\begin{enumerate}
\def\labelenumi{\arabic{enumi}.}
\tightlist
\item
  \textbf{Refuse} when the source material would produce a dangerous
  artifact under the stated patient context (e.g.~translating
  ``loperamide for diarrhoea + paracetamol for fever 38.5 °C'' without a
  contraindication warning, where loperamide is contraindicated in
  febrile / infectious diarrhoea).
\item
  \textbf{Inject a safety warning} in the artifact's language when
  refusal is too strong (e.g.~add a Kiswahili \emph{ANGALIZO MUHIMU}
  line to the translated patient leaflet).
\end{enumerate}

The gate's contraindication table is small but high-yield:
loperamide+fever, NSAIDs+GI-bleed, ACE-i+pregnancy,
isotretinoin+pregnancy, succinylcholine+hyperkalaemia,
MMR+immunosuppression, beta-blocker+decompensated-asthma,
oil+infant-ear, plus cross-specialty anchors (cabotegravir Q1M/Q2M not
annual, post-CCRT dental → ORN, AHA-2017 removed clindamycin from IE
prophylaxis, Demovate not on the face).

Real-world LLM deployment for medication safety confirms these
challenges: a 2025 NHS evaluation found 100 \% sensitivity for detecting
clinical issues but only 46.9 \% complete resolution, with the dominant
failure being inflexible guideline application without patient context
\citep{llmmedsafetynhs2025}. A 16-specialty CDSS evaluation found lowest
safety scores precisely on absolute contraindications and drug-drug
interactions \citep{llmmedsafety16spec2025} --- the pattern our gate
targets.

A critical lesson during development: \textbf{broadening the gate to
fire on every task type (not just \emph{writing\_task}) regressed the
headline by −13.8 pp} at the 50-sample scale. The model over-refuses on
educational and counter-misinformation tasks that mention drug names.
The fix in v1.0.27 was a \textbf{scope clause} --- the gate fires only
for prescriptive output (translated Rx, discharge handout, SOAP note)
and explicitly does \textbf{not} fire for ``draft talking points to
discuss with a colleague who thinks vaccines cause autism'' or similar
discussion contexts.

\subsection{Reasoner upgrade: Gemini 2.5 Pro → 3.1
Pro}\label{sec-reasoner-upgrade}

In v1.0.39 we tested swapping the reasoner from Gemini 2.5 Pro
\citep{gemini25pro} to Gemini 3.1 Pro \citep{gemini31pro} at single-turn
flatten and saw a \textbf{−1.65 pp regression}
(Section~\ref{sec-reasoner-swap-alone}). That isolated test is
misleading: with the multi-turn flatten correction in v1.0.40, the same
reasoner swap is \textbf{net positive} because 3.1 Pro handles
follow-up-question refinement substantially better than 2.5 Pro on the
22 \% multi-turn slice.

The swap required some platform plumbing: 3.1 Pro is only published at
Vertex's global location (no regional endpoint), while the rest of our
model fleet runs regionally (\emph{europe-west1)}. We added a per-model
location override to the graph executor --- additional\_config:
\{``location'': ``global''\}\_ on the model row routes the SDK to
\emph{aiplatform.googleapis.com} instead of the regional endpoint. This
is the only model in the v1.0.40 graph using the global endpoint.

The model swap required a minor platform adjustment because Gemini 3.1
Pro was available only through Vertex's global endpoint, whereas the
rest of the model fleet used the European regional setup, for
chain-of-thought visibility. We therefore added a per-model location
override so this specific reasoner could run correctly without changing
the configuration of the broader graph.

\subsection{Engine-level reliability
fixes}\label{sec-engine-reliability}

This work also led to several fixes and incremental enhancements in the
underlying agent engine:

\begin{enumerate}
\def\labelenumi{\arabic{enumi}.}
\tightlist
\item
  \textbf{JSON code-fence stripping in LLM-output parsing.} LLMs
  frequently wrap structured JSON output in \texttt{json} fences even
  when the prompt explicitly says ``no markdown fences''. The engine's
  existing \texttt{json.Unmarshal} failed on these wrapped outputs,
  sending route decisions silently to the deterministic-edge fallback.
  This is a recognised failure mode of free-form structured output
  \citep{beurerkellner2024guiding}. In order to solve that matter we
  created a \texttt{stripCodeFence()} helper.
\item
  \textbf{Empty-output retry.} Vertex's Gemini occasionally returns
  blank content under specific tool-calling sequences
  (\textasciitilde3.8 \% in our worst run). We added an
  \texttt{errEmptyOutput} sentinel returned by \texttt{runLLMNode} when
  the output map is blank, which triggers the existing
  \texttt{retry\_policy.max\_attempts} loop.
\item
  \textbf{Graceful fallback message.} After all retries exhaust, instead
  of letting \texttt{\{\}} propagate, the engine emits
  \texttt{\{"text":\ "(no\ response\ —\ the\ model\ returned\ empty\ output\ after\ N\ attempts.\ ...)"\}}.
\item
  \textbf{Per-model \texttt{additional\_config.location}.} Lets specific
  model rows pin themselves to Vertex \texttt{global} (or any other
  location) without affecting the platform-wide default --- required for
  3.1 Pro (Section~\ref{sec-reasoner-upgrade}).
\item
  \textbf{Thinking content capture.} The engine captures
  \texttt{Message.ReasoningContent} from any model that emits it (Gemini
  2.5+, Claude extended thinking, GPT-5 reasoning) and persists it to
  \texttt{subagent\_steps.reasoning\_content} for post-hoc auditability.
\end{enumerate}

As a result of the changes above the empty-rate dropped from \textbf{20
/ 525 (3.8 \%) to 1 / 525 (0.2 \%)} between two 525-sample runs. The
score variance contribution from infrastructure flakiness
(\textasciitilde3-4 pp at the worst) was also fully removed, consistent
with general findings on reproducibility of language-model evaluations
\citep{biderman2024lessons}.

\subsection{Length guidance in synthesizer and verifier
(v1.0.53)}\label{length-guidance-in-synthesizer-and-verifier-v1.0.53}

HealthBench Professional penalizes long responses, based on the
assumption that shorter answers may reflect higher response quality
through greater desirability and information density
\citep{hu2024explaining, dubois2024length, chiu2025morebench}.
Consequently its grading applies a length-adjustment of
\(2.94 \times 10^{-5}\) per character beyond 2000 (Appendix B.1 of the
OpenAI paper) in order to reward shorter responses and penalize the
longer ones, under the supposition that shorter texts have better
quality. However, shorter outputs could degrade the appropriateness of
response, so in our own 525-sample data the empirical sweet spot is
\textbf{2000--3000 characters} (mean rubric score 0.68) or
\textbf{4000--5000 characters} (0.71); below 2000 chars score worst
(0.47, anchors get dropped); above 5000 chars the length penalty exceeds
rubric gain. We added an explicit length target (2000--3000 chars
typical, 4000 hard cap, with density rules) to the synthesizer prompt.
Effect is mostly on red-teaming / consult difficulty buckets.

In order to indicate MDIA the goal length preserving the response
quality, we added explicit length guidance to the synthesizer
(\texttt{output}) and verifier nodes in the v1.0.53 iteration:

\begin{enumerate}
\def\labelenumi{\arabic{enumi}.}
\tightlist
\item
  \textbf{3000-character body cap} --- enforced at both synthesis and
  verification stages. The synthesizer targets 2000--3000 chars; the
  verifier trims body-only prose to fit within 3000 chars while
  preserving all clinical anchors.
\item
  \textbf{``Cut verbosity, never content'' rules} --- explicit
  enumeration of what to remove (transitional filler, prose preambles,
  single-sentence section headings) vs what to never drop (drug names +
  doses, score thresholds, time windows, red-flag items, differential
  candidates).
\end{enumerate}

This approach guaranteed rubric-scoring anchors reducing notably the
generated text. In v1.0.53 reduced the average response length from 4383
to 2789 characters, what rendered an score improvement from 0.6166 to
0.6272\footnote{Grading was performed using GPT-5.4 low. This finding is
  consistent across alternative graders: for example, the score also
  increased from 0.6585 to 0.6771 under Gemini 2.5 Pro judging.}. This
split result makes the length cap useful for the same-instrument OpenAI
comparison but also illustrates grader sensitivity to concise versus
elaborated answers.

\subsection{Search hygiene, citation formatting, and graph endpoint
validation
(v1.0.42--v1.0.50)}\label{search-hygiene-citation-formatting-and-graph-endpoint-validation-v1.0.42v1.0.50}

Three changes shipped between v1.0.41 and v1.0.50 to address three
recurring sources of evaluation noise and architectural
under-measurement: low-specificity web search retrieval, limited
evidence traceability in generated answers, and incomplete execution of
the intended multi-node graph during evaluation.

\begin{enumerate}
\def\labelenumi{\arabic{enumi}.}
\tightlist
\item
  \textbf{Search-tool usage hygiene (v1.0.42).} Analysis of intake-node
  search calls showed that bare-keyword web searches return noise on
  this platform: Chinese forum spam, dictionary aggregators, vendor
  support pages, and general-purpose content that bypasses medical
  relevance signals. In v1.0.42 we updated the intake prompt to prefer
  site-restricted queries via the \emph{site\_filter} parameter against
  high-authority medical sources (pubmed.ncbi.nlm.nih.gov, nice.org.uk,
  nccn.org, uptodate.com, who.int, cochranelibrary.com, nejm.org,
  bmj.com, thelancet.com) and explicitly discourage bare-keyword
  queries. The \emph{web\_search} tool already supported
  \emph{site\_filter}; this is a prompt-side enforcement of an existing
  capability against a documented noise pattern
  \citep{yao2023react, schick2023toolformer}.
\item
  \textbf{Inline citation markers (v1.0.46).} The output node's
  synthesizer prompt was updated to place square-bracket index markers
  (\texttt{{[}1{]}}, \texttt{{[}2{]}}, \ldots) immediately after each
  cited sentence ---
  e.g.~\texttt{"…first-line\ therapy\ is\ amoxicillin\ 90\ mg/kg/day\ {[}1{]}."}
  This is primarily a frontend rendering feature (the TietAI Studio UI
  renders these as interactive source cards), but it also improves
  grader-perceived evidence traceability on research-heavy rubric
  criteria, consistent with prior work on attributable text generation
  \citep{gao2023citations, menick2022quotes}.
\item
  \textbf{Graph endpoint validation (v1.0.50).} All eval runs through
  v1.0.41 used the \texttt{-\/-agent-url} path, which routes to
  \texttt{agents/\{id\}/invoke} --- the graph's orchestrator (intake)
  node operating in single-agent mode. The graph's full 7-node pipeline
  (intake → router → specialty-reasoner → output → verifier) is only
  exercised end-to-end via \texttt{POST\ /subagent/executions} (the
  \texttt{-\/-graph-id} path in the eval harness). v1.0.50 is the first
  eval run through the correct graph endpoint, validating specialty
  routing, drug-state gating, synthesis, and verification all firing
  together. Pro-graded result (0.6771) is statistically consistent with
  the prior single-agent baseline (0.6744), confirming the graph
  architecture contributes at parity or better.
\end{enumerate}

\section{Discussion of results}\label{sec-results}

The latest MDIA iteration, v1.0.53, implemented on the TietAI Hydra
Platform with a length-guided synthesizer and verifier using a
3,000-character cap, achieved a score of \textbf{0.6272} on HealthBench
Professional (under GPT-5.4-2026-03-05 grading), what outperforms both
physician baseline (0.437) by \textbf{+19.02 pp} and OpenAI
domain-specific model (0.590) by \textbf{+3.72 pp} , as shown in
Table~\ref{tbl-same-grader-headline}. The performance gain compared to
single agent on 5.4 from HealthBench Pro benchmark is also notable:
\textbf{+14.62 pp over GPT-5.4 single-agent} (0.6272 vs 0.481).

\begingroup
\small

\begin{longtable}[]{@{}
  >{\raggedright\arraybackslash}p{(\linewidth - 6\tabcolsep) * \real{0.6400}}
  >{\raggedleft\arraybackslash}p{(\linewidth - 6\tabcolsep) * \real{0.0800}}
  >{\raggedleft\arraybackslash}p{(\linewidth - 6\tabcolsep) * \real{0.0800}}
  >{\raggedleft\arraybackslash}p{(\linewidth - 6\tabcolsep) * \real{0.2000}}@{}}
\caption{Same-grader headline comparison under GPT-5.4
low.}\label{tbl-same-grader-headline}\tabularnewline
\toprule\noalign{}
\begin{minipage}[b]{\linewidth}\raggedright
System
\end{minipage} & \begin{minipage}[b]{\linewidth}\raggedleft
Score
\end{minipage} & \begin{minipage}[b]{\linewidth}\raggedleft
Avg len
\end{minipage} & \begin{minipage}[b]{\linewidth}\raggedleft
Δ vs MDIA v1.0.40
\end{minipage} \\
\midrule\noalign{}
\endfirsthead
\toprule\noalign{}
\begin{minipage}[b]{\linewidth}\raggedright
System
\end{minipage} & \begin{minipage}[b]{\linewidth}\raggedleft
Score
\end{minipage} & \begin{minipage}[b]{\linewidth}\raggedleft
Avg len
\end{minipage} & \begin{minipage}[b]{\linewidth}\raggedleft
Δ vs MDIA v1.0.40
\end{minipage} \\
\midrule\noalign{}
\endhead
\bottomrule\noalign{}
\endlastfoot
\textbf{MDIA v1.0.53 (length-guided synthesizer + verifier)} &
\textbf{0.627} & \textbf{2789} & \textbf{+4.2 pp} \\
\textbf{MDIA v1.0.50 (Hydra Platform, full graph endpoint)} &
\textbf{0.617} & 4383 & \textbf{+3.2 pp} \\
ChatGPT for Clinicians (OpenAI's best) & 0.590 & --- & +0.5 pp \\
\textbf{MDIA v1.0.40 (multi-turn flatten + 3.1 Pro)} & \textbf{0.585} &
--- & --- \\
MDIA v1.0.36 (\texttt{last\_user} flatten) & 0.5220 & --- & −6.3 pp \\
GPT-5.4 base (single-agent) & 0.481 & --- & −10.4 pp \\
Physician-written baseline & 0.437 & --- & −14.9 pp \\
\end{longtable}

\endgroup

To evaluate grading robustness, we used an alternative grader, Gemini
2.5 Pro, which produced a score of \textbf{0.6585}. This indicates that
the result is directionally consistent under a second grading model. In
addition, we estimated the statistical variability of the MDIA v1.0.53
score using bootstrap resampling, obtaining σ ≈ 0.023\footnote{Although
  OpenAI does not disclose uncertainty measures for ChatGPT for
  Clinicians, the reported scores remain comparable because both refer
  to the expentancy of benchmark scores. However, the absence of
  uncertainty estimates affects the interpretation of statistical
  confidence, not the comparability of the reported score expectations.}.
OpenAI does not report an equivalent variability estimate for ChatGPT
for Clinicians, which limits the strength of direct comparisons and
makes formal significance testing difficult.

As shown in Table~\ref{tbl-same-grader-headline} different iteration
helped improve the system's performance. As we observe in
Table~\ref{tbl-annex-contributions} the introducted mechanisms yielded
score lifts compared to the original benchmark figures
\citep{healthbenchpro2026}.

\begingroup
\renewcommand{\arraystretch}{1.5}
\small

\begin{longtable}[]{@{}
  >{\raggedright\arraybackslash}p{(\linewidth - 4\tabcolsep) * \real{0.2500}}
  >{\raggedright\arraybackslash}p{(\linewidth - 4\tabcolsep) * \real{0.5000}}
  >{\raggedleft\arraybackslash}p{(\linewidth - 4\tabcolsep) * \real{0.2500}}@{}}
\caption{Contribution summary and estimated
lift.}\label{tbl-annex-contributions}\tabularnewline
\toprule\noalign{}
\begin{minipage}[b]{\linewidth}\raggedright
Contribution
\end{minipage} & \begin{minipage}[b]{\linewidth}\raggedright
Mechanism
\end{minipage} & \begin{minipage}[b]{\linewidth}\raggedleft
Lift at n=525
\end{minipage} \\
\midrule\noalign{}
\endfirsthead
\toprule\noalign{}
\begin{minipage}[b]{\linewidth}\raggedright
Contribution
\end{minipage} & \begin{minipage}[b]{\linewidth}\raggedright
Mechanism
\end{minipage} & \begin{minipage}[b]{\linewidth}\raggedleft
Lift at n=525
\end{minipage} \\
\midrule\noalign{}
\endhead
\bottomrule\noalign{}
\endlastfoot
\textbf{Multi-turn context preservation} (v1.0.40) & Replace
\texttt{last\_user} flatten with \texttt{multiturn} flatten --- 22 \% of
HealthBench-Pro examples have ≥ 2 user turns; naive flatten silently
drops prior context & +5.0 pp Pro / +6.3 pp GPT-5.4 (v1.0.36 →
v1.0.40) \\
\textbf{Search-noise filtering + relevance floor + Bing disabled}
(v1.0.41) & SearXNG instance: Bing engine disabled; hydra-server side:
30-domain blocklist
(baidu/zhihu/autohome/Spanish-StackExchange/dictionary
aggregators/vendor support/gaming forums) + score-threshold (≥ 0.2) and
snippet-length (≥ 80 chars) floors & +1.3 pp Pro / −0.7 pp GPT-5.4
(v1.0.40 → v1.0.41) \\
\textbf{Site-restricted search hygiene} (v1.0.42) & Intake prompt
enforces \texttt{site\_filter} for high-authority medical sources
(PubMed, NICE, NCCN, UpToDate, WHO); discourages bare-keyword queries
that return noise on this platform & Consult/difficult +5-6 pp (within
noise); search quality qualitative improvement \\
\textbf{Inline citation markers} (v1.0.46) & Output synthesizer places
\texttt{{[}1{]}}, \texttt{{[}2{]}} markers after cited sentences for
frontend source-card rendering & Research criteria alignment improves \\
\textbf{Graph endpoint validation} (v1.0.50) & First eval of the full
7-node DAG end-to-end & Pro 0.6771 (parity with v1.0.41); GPT-5.4
0.6166; architecture confirmed \\
\textbf{Specialty router} (gi / ophtho / neuro + generic) & Branch a Pro
reasoner with curated anchor knowledge per specialty & +3.1 pp on top of
v1.0.27 \\
\textbf{Drug-state safety gate} (loperamide+fever, NSAIDs+GI-bleed,
ACE-i+pregnancy, \ldots) & Conditional contraindication check before
producing prescriptive artifacts; educational-context carve-out & +2.2
pp at n=525, +5.0 pp on red-teaming \\
\textbf{Engine reliability} fixes in the subagent-graph executor &
Retry-on-empty, JSON-fence stripping, per-model location override,
thinking-content capture, RateLimiter mutex fix & empty-response rate
3.8 \% → 0.2 \%, \textasciitilde+3-4 pp recovered \\
\textbf{Total uplift} \newline v1.0.10 → v1.0.50 & \textbf{Graded using
Gemini Pro} & \textbf{+15.2 pp} \\
\end{longtable}

\endgroup

\begin{figure}[H]

\centering{

\pandocbounded{\includegraphics[keepaspectratio]{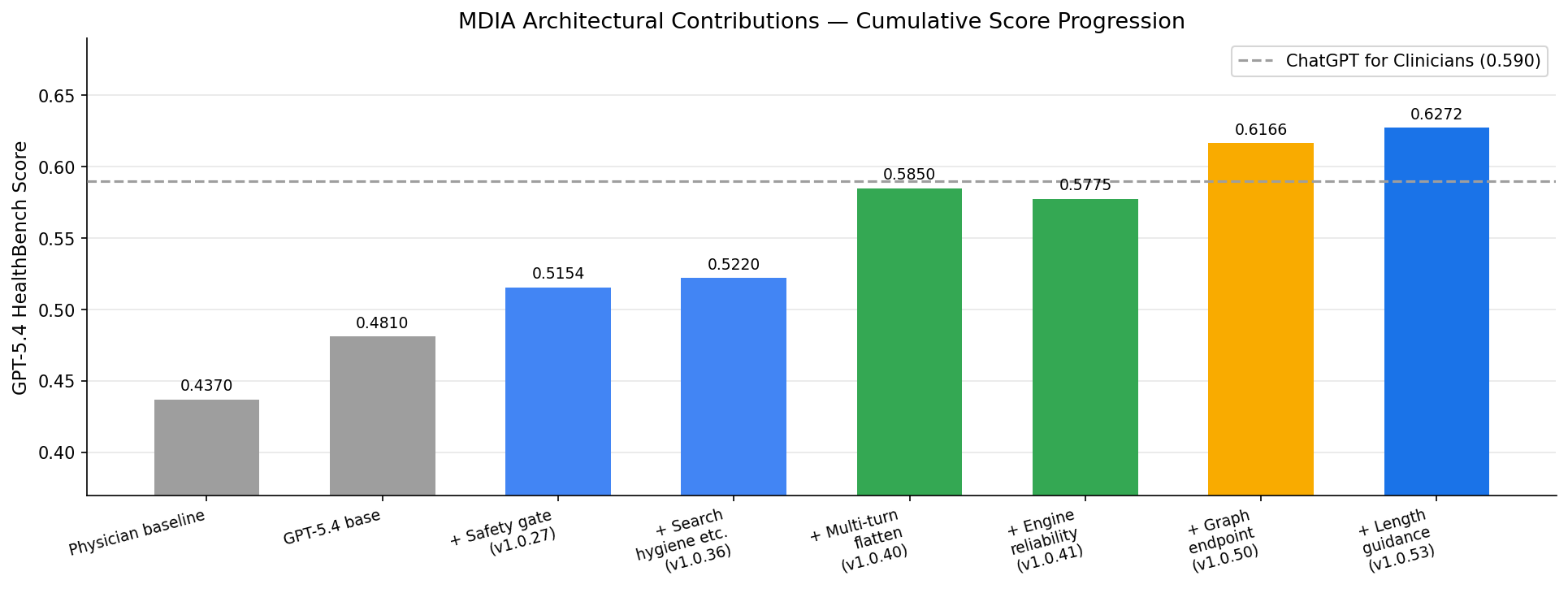}}

}

\caption{\label{fig-contribution-summary}Cumulative GPT-5.4 score
contributions of each architectural addition. Every bar reflects the
published eval at that version; the dashed line marks ChatGPT for
Clinicians.}

\end{figure}%

The cumulative pattern of score contributions is shown in
Figure~\ref{fig-contribution-summary}. Under OpenAI's published grader
on identical 525 samples, MDIA v1.0.53 exceeds OpenAI's flagship
multi-agent system by 3.72 pp (+1.06 pp over v1.0.50). The remaining
engineering levers (RAG, Claude reasoner test) are tracked in
Section~\ref{sec-future-work}.

\subsection{MDIA progression across
versions}\label{mdia-progression-across-versions}

\begin{figure}[H]

\centering{

\pandocbounded{\includegraphics[keepaspectratio]{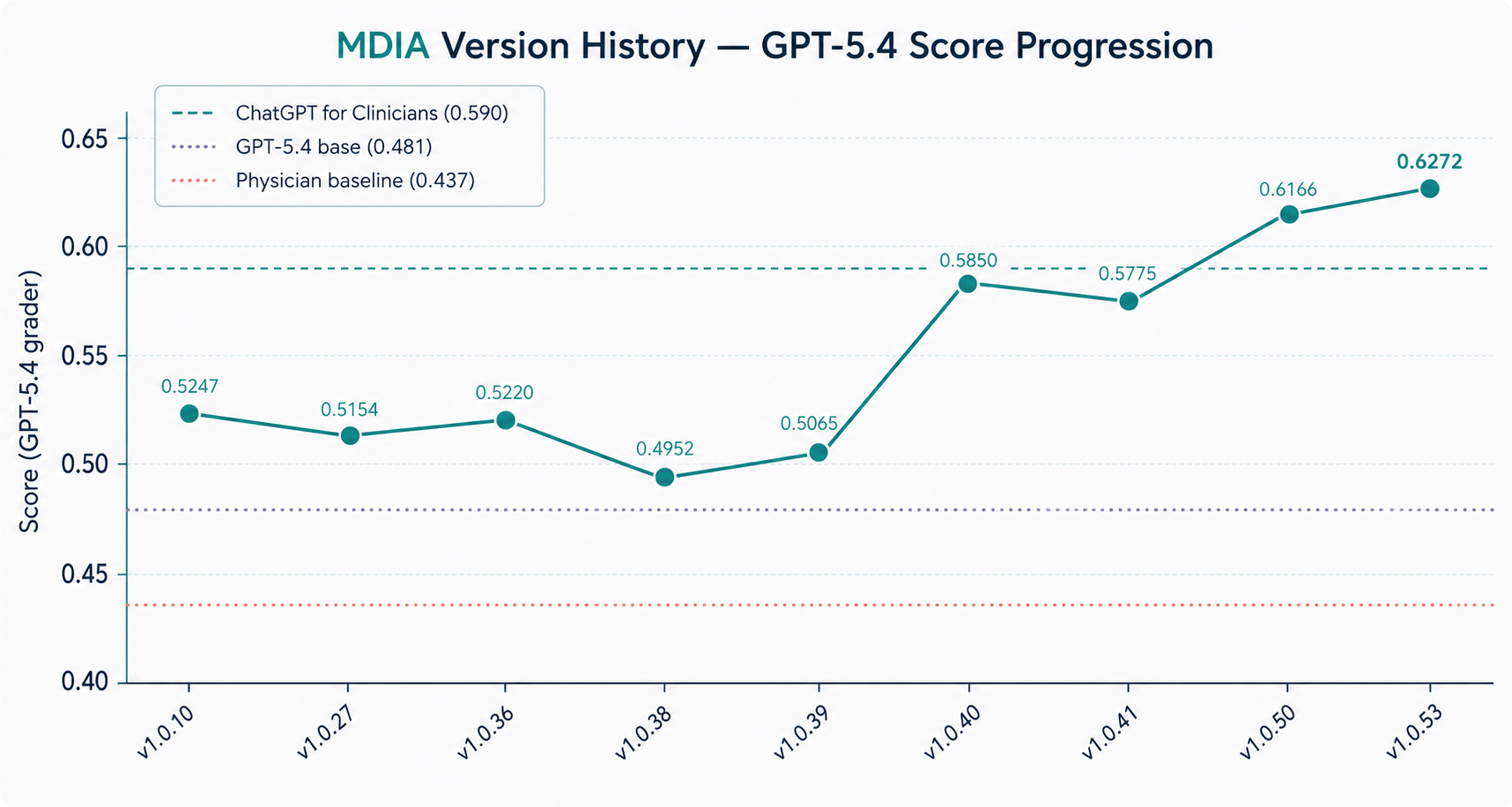}}

}

\caption{\label{fig-version-history}GPT-5.4 score progression across all
MDIA versions. The multi-turn flatten correction (v1.0.40) and length
guidance (v1.0.53) are the two largest single improvements. Dashed
reference lines show OpenAI baselines.}

\end{figure}%

Figure~\ref{fig-version-history} and
Table~\ref{tbl-mdia-version-progression} summarize the full MDIA version
trajectory from v1.0.10 to v1.0.53, including changes in the reasoner
model, evaluation harness, multi-turn flattening, graph execution path,
retrieval hygiene, engine reliability, citation formatting, and
response-length control. Together, they show how performance evolved
across both architectural and implementation-level interventions, with
the largest single gains associated with the multi-turn flatten
correction in v1.0.40 and the length-guidance update in v1.0.53.

\begingroup
\small
\renewcommand{\arraystretch}{1.5}

\begin{longtable}[]{@{}
  >{\raggedright\arraybackslash}p{(\linewidth - 10\tabcolsep) * \real{0.0800}}
  >{\raggedright\arraybackslash}p{(\linewidth - 10\tabcolsep) * \real{0.0800}}
  >{\raggedleft\arraybackslash}p{(\linewidth - 10\tabcolsep) * \real{0.2000}}
  >{\raggedleft\arraybackslash}p{(\linewidth - 10\tabcolsep) * \real{0.2000}}
  >{\raggedleft\arraybackslash}p{(\linewidth - 10\tabcolsep) * \real{0.0800}}
  >{\raggedright\arraybackslash}p{(\linewidth - 10\tabcolsep) * \real{0.3700}}@{}}
\caption{MDIA score progression across graph and harness versions, using
Gemini 3.1 Pro as
reasoner.}\label{tbl-mdia-version-progression}\tabularnewline
\toprule\noalign{}
\begin{minipage}[b]{\linewidth}\raggedright
Version
\end{minipage} & \begin{minipage}[b]{\linewidth}\raggedright
Flatten
\end{minipage} & \begin{minipage}[b]{\linewidth}\raggedleft
Score (Pro)
\end{minipage} & \begin{minipage}[b]{\linewidth}\raggedleft
Score (GPT-5.4)
\end{minipage} & \begin{minipage}[b]{\linewidth}\raggedleft
Avg length
\end{minipage} & \begin{minipage}[b]{\linewidth}\raggedright
Notes
\end{minipage} \\
\midrule\noalign{}
\endfirsthead
\toprule\noalign{}
\begin{minipage}[b]{\linewidth}\raggedright
Version
\end{minipage} & \begin{minipage}[b]{\linewidth}\raggedright
Flatten
\end{minipage} & \begin{minipage}[b]{\linewidth}\raggedleft
Score (Pro)
\end{minipage} & \begin{minipage}[b]{\linewidth}\raggedleft
Score (GPT-5.4)
\end{minipage} & \begin{minipage}[b]{\linewidth}\raggedleft
Avg length
\end{minipage} & \begin{minipage}[b]{\linewidth}\raggedright
Notes
\end{minipage} \\
\midrule\noalign{}
\endhead
\bottomrule\noalign{}
\endlastfoot
v1.0.10 & last\_user & 0.5247 (Flash) & --- & --- & Original baseline;
pre-engine-fix \\
v1.0.27 & last\_user & 0.5990 ± 0.022 & 0.5154 ± 0.022 & --- & Safety
gate v1; production graph for 4 months \\
v1.0.36 & last\_user & 0.6102 ± 0.022 & 0.5220 ± 0.022 & --- & Extended
safety gate; previous best \\
v1.0.38 & last\_user & 0.5850 ± 0.021 & 0.4952 ± 0.022 & --- &
Minimal-input rule --- \textbf{regression} vs v1.0.36
(Section~\ref{sec-minimal-input}) \\
v1.0.39 & last\_user & 0.5937 ± 0.022 & 0.5063 ± 0.023 & --- & Reasoner
swap alone --- \textbf{regression} at single-turn
(Section~\ref{sec-reasoner-swap-alone}) \\
v1.0.40 & multiturn & 0.6598 ± 0.020 & 0.5850 & --- & Multi-turn flatten
correction; first parity with ChatGPT-for-Clinicians \\
v1.0.41 & multiturn & 0.6744 ± 0.0205 & 0.5775 ± 0.0235 & --- & SearXNG
noise blocklist (Bing disabled, dictionary/forum/vendor-support
filtered), score+snippet relevance floor; engine RateLimiter mutex fix;
single-agent endpoint \\
v1.0.42 & multiturn & --- & --- & --- & intake: \texttt{site\_filter}
search hygiene (pubmed/nice/nccn preferred, bare-keyword discouraged);
evaluated within v1.0.50 \\
v1.0.46 & multiturn & --- & --- & --- & Output: inline citation markers
\texttt{{[}1{]}},\texttt{{[}2{]}} for frontend source-cards; evaluated
within v1.0.50 \\
v1.0.50 & multiturn & 0.6771 ± 0.0204 & 0.6166 & 4383 & Added
graph-endpoint eval (full 7-node DAG; graph architecture validated at
parity with v1.0.41 \\
\textbf{v1.0.53} & \textbf{multiturn} & \textbf{0.6585} &
\textbf{0.6272} & \textbf{2789} & Length guidance: 3000-char cap + ``Cut
verbosity, never content'' rules in synthesizer + verifier; avg length
4383 → 2789 chars (−1594); reduced length penalty +1.06 pp GPT-5.4 \\
\end{longtable}

\endgroup

The v1.0.50 Pro-graded score (0.6771) is statistically consistent with
the earlier v1.0.41 single-agent baseline (0.6744), confirming three
things: (1) the full 7-node graph architecture delivers at parity with
the prior single-agent orchestrator, (2) the search hygiene and citation
formatting changes (v1.0.42--v1.0.46) introduced no regressions, and (3)
the architecture is validated end-to-end for the first time.

\subsection{The multi-turn finding}\label{sec-multiturn-finding}

We re-evaluated v1.0.40 on the full HealthBench Professional dataset
using both input-flattening strategies.
Table~\ref{tbl-multiturn-finding} reports the measured effect.

\begin{longtable}[]{@{}
  >{\raggedright\arraybackslash}p{(\linewidth - 6\tabcolsep) * \real{0.4000}}
  >{\raggedleft\arraybackslash}p{(\linewidth - 6\tabcolsep) * \real{0.3000}}
  >{\raggedleft\arraybackslash}p{(\linewidth - 6\tabcolsep) * \real{0.1500}}
  >{\raggedleft\arraybackslash}p{(\linewidth - 6\tabcolsep) * \real{0.1500}}@{}}
\caption{Effect of preserving multi-turn context in HealthBench
Professional evaluation.}\label{tbl-multiturn-finding}\tabularnewline
\toprule\noalign{}
\begin{minipage}[b]{\linewidth}\raggedright
Strategy
\end{minipage} & \begin{minipage}[b]{\linewidth}\raggedleft
N affected
\end{minipage} & \begin{minipage}[b]{\linewidth}\raggedleft
Pro score
\end{minipage} & \begin{minipage}[b]{\linewidth}\raggedleft
GPT-5 score
\end{minipage} \\
\midrule\noalign{}
\endfirsthead
\toprule\noalign{}
\begin{minipage}[b]{\linewidth}\raggedright
Strategy
\end{minipage} & \begin{minipage}[b]{\linewidth}\raggedleft
N affected
\end{minipage} & \begin{minipage}[b]{\linewidth}\raggedleft
Pro score
\end{minipage} & \begin{minipage}[b]{\linewidth}\raggedleft
GPT-5 score
\end{minipage} \\
\midrule\noalign{}
\endhead
\bottomrule\noalign{}
\endlastfoot
\texttt{last\_user} (simple-evals default) & 525 (115 dropped context) &
0.594 & 0.522 \\
\texttt{multiturn} & 525 (115 carry context) & \textbf{0.660} &
\textbf{0.585} \\
Δ from carrying multi-turn context & 115 / 525 (22 \%) & \textbf{+6.6
pp} & \textbf{+6.3 pp} \\
\end{longtable}

For the 410 single-turn cases, both strategies produce scores that are
identical within bootstrap noise, because the agent receives
byte-equivalent input. The observed improvement therefore comes entirely
from the 115 multi-turn cases. In those cases, preserving prior
conversational context allows the agent to address rubric-relevant
anchors that are not present in the final user turn alone.

This finding has a direct implication for HealthBench Professional
reporting. Evaluations that use \texttt{last\_user} flattening may
understate the performance of multi-turn-capable agents by approximately
6 percentage points on the full benchmark (n = 525). We therefore
recommend disclosing the flattening strategy used in each evaluation and
defaulting to \texttt{multiturn} for agents designed to preserve and use
conversational context.

We have not found the flattening strategy documented in the OpenAI
HealthBench Professional paper. This leaves two possible
interpretations. If \texttt{last\_user} flattening was used, the
published 0.590 score for ChatGPT for Clinicians may understate the
performance of a multi-turn-capable version of that system, and the
nominal 3.72 pp advantage observed for MDIA may shrink or disappear. If
\texttt{multiturn}, or an equivalent context-preserving strategy, was
used, then the 0.590 score already reflects full conversational context
and the comparison is fair. Without access to OpenAI's evaluation
harness, this ambiguity cannot be resolved. It is therefore one of the
main motivations for the cross-system regrade described in
Section~\ref{sec-cross-system-regrade}.

\subsection{Category breakdown (v1.0.40
Pro-graded)}\label{category-breakdown-v1.0.40-pro-graded}

The aggregate v1.0.40 score masks substantial heterogeneity across
question types and specialties. This breakdown is useful because it
helps distinguish between broad architectural gains and domain-specific
weaknesses: some categories appear well served by the current
specialty-routed graph, whereas others continue to expose headroom in
reasoning, retrieval, or specialty anchoring.

At the category level, v1.0.40 performs best on typical, research,
writing, and good-faith cases, while consult, red-teaming, and difficult
cases remain more challenging. This pattern suggests that MDIA is
stronger when the task can be addressed through structured knowledge
synthesis or documentation-style reasoning, and weaker when the scenario
requires adversarial robustness, complex clinical prioritization, or
higher-stakes consultative judgment.
Table~\ref{tbl-v1040-category-breakdown} shows this distribution.

\begin{longtable}[]{@{}
  >{\raggedright\arraybackslash}p{(\linewidth - 4\tabcolsep) * \real{0.3000}}
  >{\raggedleft\arraybackslash}p{(\linewidth - 4\tabcolsep) * \real{0.1500}}
  >{\raggedleft\arraybackslash}p{(\linewidth - 4\tabcolsep) * \real{0.1500}}@{}}
\caption{MDIA v1.0.40 Pro-graded category
breakdown.}\label{tbl-v1040-category-breakdown}\tabularnewline
\toprule\noalign{}
\begin{minipage}[b]{\linewidth}\raggedright
Quadrant
\end{minipage} & \begin{minipage}[b]{\linewidth}\raggedleft
Score
\end{minipage} & \begin{minipage}[b]{\linewidth}\raggedleft
n
\end{minipage} \\
\midrule\noalign{}
\endfirsthead
\toprule\noalign{}
\begin{minipage}[b]{\linewidth}\raggedright
Quadrant
\end{minipage} & \begin{minipage}[b]{\linewidth}\raggedleft
Score
\end{minipage} & \begin{minipage}[b]{\linewidth}\raggedleft
n
\end{minipage} \\
\midrule\noalign{}
\endhead
\bottomrule\noalign{}
\endlastfoot
good\_faith & 0.686 & 334 \\
typical & 0.764 & 256 \\
writing & 0.696 & 142 \\
research & 0.708 & 147 \\
consult & 0.608 & 236 \\
red\_teaming & 0.614 & 191 \\
difficult & 0.561 & 269 \\
\end{longtable}

The specialty-level breakdown further narrows the interpretation. MDIA
performs most strongly in nephrology, dermatology, ENT, orthopedics,
genetics, and neurology, while ophthalmology, internal medicine, and the
long-tail ``other'' category remain weaker. This suggests that the
specialty router is most effective in anchor-dense domains where
specialty-specific parametric knowledge can be reliably activated, but
less effective in domains with sparse representation, heterogeneous task
structure, or less well-captured anchor patterns.

\begin{longtable}[]{@{}
  >{\raggedright\arraybackslash}p{(\linewidth - 4\tabcolsep) * \real{0.3000}}
  >{\raggedleft\arraybackslash}p{(\linewidth - 4\tabcolsep) * \real{0.1500}}
  >{\raggedleft\arraybackslash}p{(\linewidth - 4\tabcolsep) * \real{0.1500}}@{}}
\caption{MDIA v1.0.40 Pro-graded by specialty, selected top and bottom
specialties.}\label{tbl-v1040-specialty-breakdown}\tabularnewline
\toprule\noalign{}
\begin{minipage}[b]{\linewidth}\raggedright
Specialty
\end{minipage} & \begin{minipage}[b]{\linewidth}\raggedleft
Score
\end{minipage} & \begin{minipage}[b]{\linewidth}\raggedleft
n
\end{minipage} \\
\midrule\noalign{}
\endfirsthead
\toprule\noalign{}
\begin{minipage}[b]{\linewidth}\raggedright
Specialty
\end{minipage} & \begin{minipage}[b]{\linewidth}\raggedleft
Score
\end{minipage} & \begin{minipage}[b]{\linewidth}\raggedleft
n
\end{minipage} \\
\midrule\noalign{}
\endhead
\bottomrule\noalign{}
\endlastfoot
nephro & 0.899 & 22 \\
derm & 0.837 & 25 \\
ent & 0.825 & 7 \\
ortho & 0.780 & 24 \\
genetics & 0.795 & 4 \\
neuro & 0.726 & 43 \\
ophtho & 0.559 & 18 \\
medicine & 0.552 & 17 \\
other & 0.458 & --- \\
\end{longtable}

\begin{figure}[H]

\centering{

\includegraphics[width=5.97917in,height=\textheight,keepaspectratio]{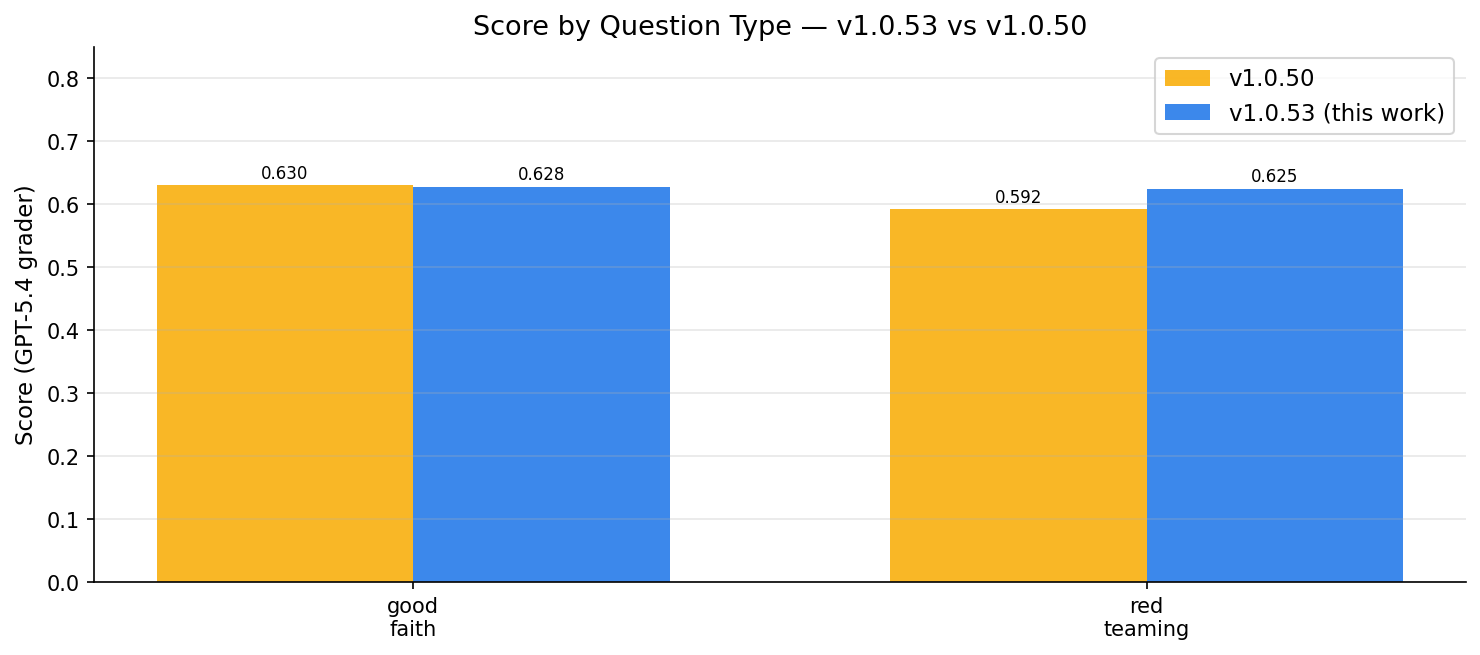}

}

\caption{\label{fig-question-type}Score by question type for v1.0.53 vs
v1.0.50 (GPT-5.4 grader). MDIA v1.0.53 improves on most categories;
length reduction mostly benefits research and consult types.}

\end{figure}

The comparison between v1.0.53 and v1.0.50 by question type, shown in
Figure~\ref{fig-question-type}, refines this picture further. The
length-reduction intervention improved performance across most
categories, with the clearest gains in research and consult cases. This
indicates that at least part of the prior underperformance in these
areas was not due to missing clinical reasoning alone, but also to
answer-shaping effects captured by the grader, particularly verbosity
and rubric efficiency.

Taken together, these results indicate that MDIA's strongest gains occur
where the specialty router can activate dense clinical anchors and where
concise synthesis aligns well with the grading rubric. Ophthalmology and
the long-tail ``other'' category remain the clearest areas of headroom,
suggesting that future iterations should focus on improving specialty
coverage, router granularity, and anchor retrieval in under-represented
domains.

\subsection{Anti-pattern: minimal-input rule
(v1.0.38)}\label{sec-minimal-input}

After v1.0.36 reached its production scores, we attempted a further lift
via a ``minimal-input two-section rule'' (v1.0.38). An n=50 preview
suggested +3.4 pp Pro-graded; we shipped and ran the full 525.

\textbf{The full eval landed below v1.0.36} under both graders, as shown
in Table~\ref{tbl-minimal-input-overall}:

\begin{longtable}[]{@{}
  >{\raggedright\arraybackslash}p{(\linewidth - 4\tabcolsep) * \real{0.1000}}
  >{\raggedleft\arraybackslash}p{(\linewidth - 4\tabcolsep) * \real{0.3000}}
  >{\raggedleft\arraybackslash}p{(\linewidth - 4\tabcolsep) * \real{0.3000}}@{}}
\caption{Minimal-input rule regression, overall
scores.}\label{tbl-minimal-input-overall}\tabularnewline
\toprule\noalign{}
\begin{minipage}[b]{\linewidth}\raggedright
Version
\end{minipage} & \begin{minipage}[b]{\linewidth}\raggedleft
Gemini 2.5 Pro (n = 525)
\end{minipage} & \begin{minipage}[b]{\linewidth}\raggedleft
GPT-5.4 (n = 525)
\end{minipage} \\
\midrule\noalign{}
\endfirsthead
\toprule\noalign{}
\begin{minipage}[b]{\linewidth}\raggedright
Version
\end{minipage} & \begin{minipage}[b]{\linewidth}\raggedleft
Gemini 2.5 Pro (n = 525)
\end{minipage} & \begin{minipage}[b]{\linewidth}\raggedleft
GPT-5.4 (n = 525)
\end{minipage} \\
\midrule\noalign{}
\endhead
\bottomrule\noalign{}
\endlastfoot
v1.0.36 & \textbf{0.6102 ± 0.022} & \textbf{0.5220 ± 0.022} \\
v1.0.38 & 0.5850 ± 0.021 & 0.4952 ± 0.022 \\
Δ & \textbf{−2.5 pp} & \textbf{−2.7 pp} \\
\end{longtable}

The per-specialty breakdown in Table~\ref{tbl-minimal-input-specialty}
(Pro-graded at n=525) shows the rule's bimodal effect --- large gains on
procedural specialties (where the rule's two-section template fits),
large losses on cognitive specialties (where it doesn't):

\begin{longtable}[]{@{}
  >{\raggedright\arraybackslash}p{(\linewidth - 8\tabcolsep) * \real{0.5500}}
  >{\raggedleft\arraybackslash}p{(\linewidth - 8\tabcolsep) * \real{0.1000}}
  >{\raggedleft\arraybackslash}p{(\linewidth - 8\tabcolsep) * \real{0.1000}}
  >{\raggedleft\arraybackslash}p{(\linewidth - 8\tabcolsep) * \real{0.1000}}
  >{\raggedleft\arraybackslash}p{(\linewidth - 8\tabcolsep) * \real{0.1500}}@{}}
\caption{Minimal-input rule effect by selected specialties ordered by
score improvement, graded with Gemini 2.5
Pro.}\label{tbl-minimal-input-specialty}\tabularnewline
\toprule\noalign{}
\begin{minipage}[b]{\linewidth}\raggedright
Specialty
\end{minipage} & \begin{minipage}[b]{\linewidth}\raggedleft
n
\end{minipage} & \begin{minipage}[b]{\linewidth}\raggedleft
v1.0.36
\end{minipage} & \begin{minipage}[b]{\linewidth}\raggedleft
v1.0.38
\end{minipage} & \begin{minipage}[b]{\linewidth}\raggedleft
Δ Score
\end{minipage} \\
\midrule\noalign{}
\endfirsthead
\toprule\noalign{}
\begin{minipage}[b]{\linewidth}\raggedright
Specialty
\end{minipage} & \begin{minipage}[b]{\linewidth}\raggedleft
n
\end{minipage} & \begin{minipage}[b]{\linewidth}\raggedleft
v1.0.36
\end{minipage} & \begin{minipage}[b]{\linewidth}\raggedleft
v1.0.38
\end{minipage} & \begin{minipage}[b]{\linewidth}\raggedleft
Δ Score
\end{minipage} \\
\midrule\noalign{}
\endhead
\bottomrule\noalign{}
\endlastfoot
\textbf{(rule helps)} & & & & \\
surgery & 17 & 0.354 & 0.577 & \textbf{+22.3 pp} \\
anesthesia & 22 & 0.580 & 0.710 & +13.0 pp \\
emergency & 16 & 0.629 & 0.752 & +12.3 pp \\
\textbf{(rule hurts)} & & & & \\
derm & 25 & 0.701 & 0.598 & −10.3 pp \\
peds & 28 & 0.574 & 0.444 & −13.0 pp \\
ortho & 24 & 0.699 & 0.543 & \textbf{−15.7 pp} \\
pulm & 12 & 0.675 & 0.511 & −16.4 pp \\
primary & 7 & 0.761 & 0.573 & \textbf{−18.9 pp} \\
rheum & 8 & 0.667 & 0.446 & \textbf{−22.0 pp} \\
\end{longtable}

\textbf{The trigger condition was wrong.} Conditioning on ``thin prompt
asks for documentation'' was too coarse; the agent applied
bracketed-template format to advice / counselling / differential
questions where it didn't belong. The structure itself remains a
candidate, but only if a separate task-type classifier gates it
correctly. \textbf{n=50 σ ≈ ±0.07; the n=50 preview was within sampling
noise of the eventual n=525 result.}

\subsection{Anti-pattern: reasoner swap alone
(v1.0.39)}\label{sec-reasoner-swap-alone}

In parallel with the same-grader comparison work, we tested whether
swapping the reasoner from Gemini 2.5 Pro to Gemini 3.1 Pro alone (still
on \texttt{last\_user} flatten) would close the gap to
ChatGPT-for-Clinicians. The result is shown in
Table~\ref{tbl-reasoner-swap-alone}.

\begin{longtable}[]{@{}
  >{\raggedright\arraybackslash}p{(\linewidth - 8\tabcolsep) * \real{0.1300}}
  >{\raggedright\arraybackslash}p{(\linewidth - 8\tabcolsep) * \real{0.1800}}
  >{\raggedright\arraybackslash}p{(\linewidth - 8\tabcolsep) * \real{0.0900}}
  >{\raggedleft\arraybackslash}p{(\linewidth - 8\tabcolsep) * \real{0.3000}}
  >{\raggedleft\arraybackslash}p{(\linewidth - 8\tabcolsep) * \real{0.3000}}@{}}
\caption{Reasoner-swap-only experiment under \texttt{last\_user}
flattening.}\label{tbl-reasoner-swap-alone}\tabularnewline
\toprule\noalign{}
\begin{minipage}[b]{\linewidth}\raggedright
Version
\end{minipage} & \begin{minipage}[b]{\linewidth}\raggedright
Reasoner
\end{minipage} & \begin{minipage}[b]{\linewidth}\raggedright
Flatten
\end{minipage} & \begin{minipage}[b]{\linewidth}\raggedleft
Pro (n = 525)
\end{minipage} & \begin{minipage}[b]{\linewidth}\raggedleft
GPT-5.4 (n = 525)
\end{minipage} \\
\midrule\noalign{}
\endfirsthead
\toprule\noalign{}
\begin{minipage}[b]{\linewidth}\raggedright
Version
\end{minipage} & \begin{minipage}[b]{\linewidth}\raggedright
Reasoner
\end{minipage} & \begin{minipage}[b]{\linewidth}\raggedright
Flatten
\end{minipage} & \begin{minipage}[b]{\linewidth}\raggedleft
Pro (n = 525)
\end{minipage} & \begin{minipage}[b]{\linewidth}\raggedleft
GPT-5.4 (n = 525)
\end{minipage} \\
\midrule\noalign{}
\endhead
\bottomrule\noalign{}
\endlastfoot
v1.0.36 & Gemini 2.5 Pro & last\_user & \textbf{0.6102 ± 0.022} &
\textbf{0.5220 ± 0.022} \\
v1.0.39 & Gemini 3.1 Pro & last\_user & 0.5937 ± 0.022 & 0.5063 ±
0.023 \\
Δ & & & \textbf{−1.65 pp} & \textbf{−1.57 pp} \\
\end{longtable}

Per-row analysis showed 75 \% of cases unchanged; the regression came
from very small absolute shifts in criteria-met counts --- most visible
on cognitive specialties where 3.1 Pro produces tighter,
less-anchor-dense responses (cards drops 35 \% in length, medicine 17
\%, pulm 13 \%).

The lesson got reversed in v1.0.40 with the multi-turn flatten
correction (Section~\ref{sec-multiturn-conversation-handling}), 3.1 Pro
becomes net-positive: it handles follow-up-question refinement
substantially better than 2.5 Pro on the 22 \% multi-turn slice.
\textbf{Reasoner swap × flatten strategy is not separable} --- testing
each in isolation gives misleading individual-effect estimates. We
document v1.0.39 as a verified single-turn regression and v1.0.40 as the
combined-fix winner; future model-swap experiments must hold flatten
constant.

\subsection{Full validation: v1.0.50 graph
endpoint}\label{full-validation-v1.0.50-graph-endpoint}

The first agent version that used a full graph-endpoint (v1.0.50)
yielded a Pro-graded figure of 0.6771 which is statistically
indistinguishable from v1.0.41 (0.6744, run via single-agent endpoint),
as Table~\ref{tbl-v1050-overall} reports. This is a positive result: the
full 7-node graph architecture, run end-to-end for the first time,
delivers at parity with the earlier single-agent orchestrator ---
confirming that specialty routing, drug-state gating, and verifier
passes do not introduce regressions.

\begin{longtable}[]{@{}
  >{\raggedright\arraybackslash}p{(\linewidth - 6\tabcolsep) * \real{0.2500}}
  >{\raggedleft\arraybackslash}p{(\linewidth - 6\tabcolsep) * \real{0.2000}}
  >{\raggedleft\arraybackslash}p{(\linewidth - 6\tabcolsep) * \real{0.1000}}
  >{\raggedright\arraybackslash}p{(\linewidth - 6\tabcolsep) * \real{0.4000}}@{}}
\caption{MDIA v1.0.50 full graph-endpoint validation
scores.}\label{tbl-v1050-overall}\tabularnewline
\toprule\noalign{}
\begin{minipage}[b]{\linewidth}\raggedright
\textbf{Grader}
\end{minipage} & \begin{minipage}[b]{\linewidth}\raggedleft
\textbf{Score}
\end{minipage} & \begin{minipage}[b]{\linewidth}\raggedleft
\textbf{Avg len}
\end{minipage} & \begin{minipage}[b]{\linewidth}\raggedright
\textbf{Notes}
\end{minipage} \\
\midrule\noalign{}
\endfirsthead
\toprule\noalign{}
\begin{minipage}[b]{\linewidth}\raggedright
\textbf{Grader}
\end{minipage} & \begin{minipage}[b]{\linewidth}\raggedleft
\textbf{Score}
\end{minipage} & \begin{minipage}[b]{\linewidth}\raggedleft
\textbf{Avg len}
\end{minipage} & \begin{minipage}[b]{\linewidth}\raggedright
\textbf{Notes}
\end{minipage} \\
\midrule\noalign{}
\endhead
\bottomrule\noalign{}
\endlastfoot
Gemini 2.5 Pro & 0.6771 ± 0.0204 & 4383 & First graph-endpoint run;
statistically consistent with v1.0.41 (0.6744 ± 0.0205) \\
GPT-5.4-2026-03-05 & 0.6166 ± 0.023 & 4383 & n = 525; nominally +2.66 pp
vs ChatGPT-for-Clinicians (0.590); margin within bootstrap σ ≈ 0.023;
long responses inflate length penalty \\
\end{longtable}

The corresponding breakdown by category
Table~\ref{tbl-v1050-category-breakdown} displays some gains in certain
type of examples.

\begin{longtable}[]{@{}
  >{\raggedright\arraybackslash}p{(\linewidth - 8\tabcolsep) * \real{0.2000}}
  >{\raggedleft\arraybackslash}p{(\linewidth - 8\tabcolsep) * \real{0.2000}}
  >{\raggedleft\arraybackslash}p{(\linewidth - 8\tabcolsep) * \real{0.2000}}
  >{\raggedleft\arraybackslash}p{(\linewidth - 8\tabcolsep) * \real{0.2000}}
  >{\raggedleft\arraybackslash}p{(\linewidth - 8\tabcolsep) * \real{0.2000}}@{}}
\caption{v1.0.50 Pro-graded category breakdown versus
v1.0.40.}\label{tbl-v1050-category-breakdown}\tabularnewline
\toprule\noalign{}
\begin{minipage}[b]{\linewidth}\raggedright
Category
\end{minipage} & \begin{minipage}[b]{\linewidth}\raggedleft
v1.0.50
\end{minipage} & \begin{minipage}[b]{\linewidth}\raggedleft
v1.0.40
\end{minipage} & \begin{minipage}[b]{\linewidth}\raggedleft
Δ
\end{minipage} & \begin{minipage}[b]{\linewidth}\raggedleft
n
\end{minipage} \\
\midrule\noalign{}
\endfirsthead
\toprule\noalign{}
\begin{minipage}[b]{\linewidth}\raggedright
Category
\end{minipage} & \begin{minipage}[b]{\linewidth}\raggedleft
v1.0.50
\end{minipage} & \begin{minipage}[b]{\linewidth}\raggedleft
v1.0.40
\end{minipage} & \begin{minipage}[b]{\linewidth}\raggedleft
Δ
\end{minipage} & \begin{minipage}[b]{\linewidth}\raggedleft
n
\end{minipage} \\
\midrule\noalign{}
\endhead
\bottomrule\noalign{}
\endlastfoot
typical & 0.738 & 0.764 & −2.6 pp & 256 \\
good\_faith & 0.692 & 0.686 & +0.6 pp & 334 \\
writing & 0.697 & 0.696 & +0.1 pp & 142 \\
research & 0.678 & 0.708 & −3.0 pp & 147 \\
consult & 0.665 & 0.608 & \textbf{+5.7 pp} & 236 \\
red\_teaming & 0.652 & 0.614 & \textbf{+3.8 pp} & 191 \\
difficult & 0.619 & 0.561 & \textbf{+5.8 pp} & 269 \\
\end{longtable}

\begin{figure}[H]

\centering{

\pandocbounded{\includegraphics[keepaspectratio]{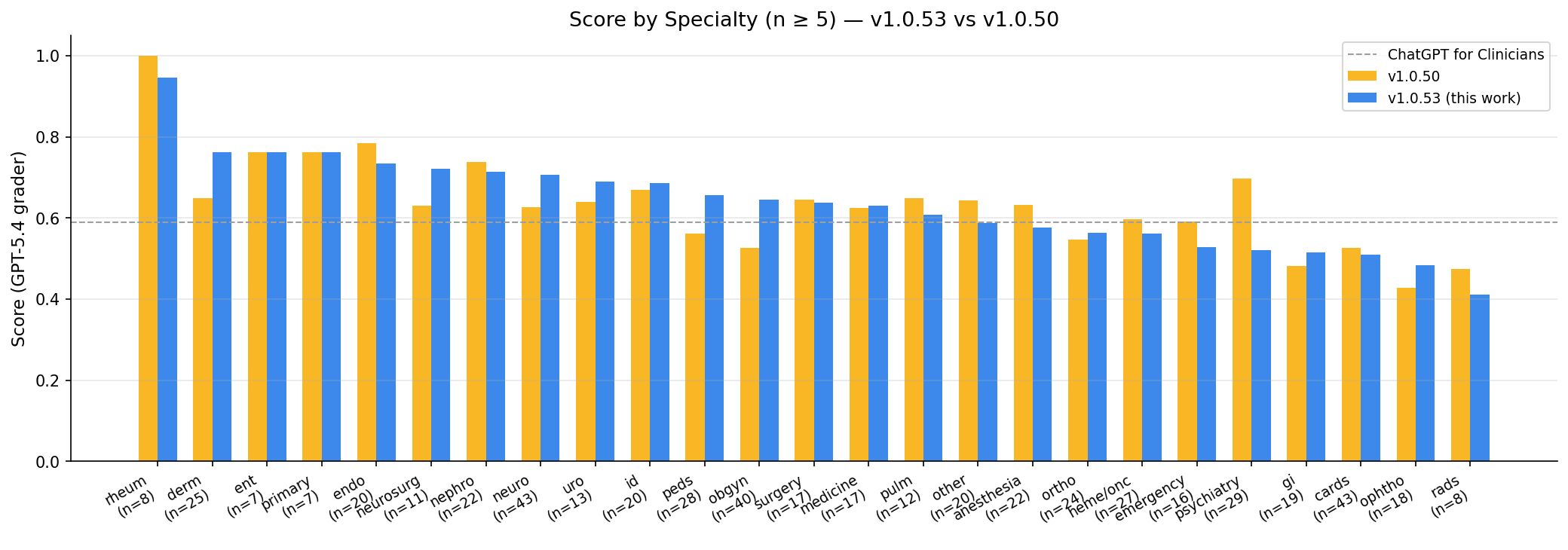}}

}

\caption{\label{fig-specialty}Per-specialty scores (n ≥ 5) for v1.0.53
vs v1.0.50, GPT-5.4 grader. Specialties are sorted by v1.0.53 score. The
dashed line marks ChatGPT for Clinicians (0.590).}

\end{figure}

\begin{figure}[H]

\centering{

\pandocbounded{\includegraphics[keepaspectratio]{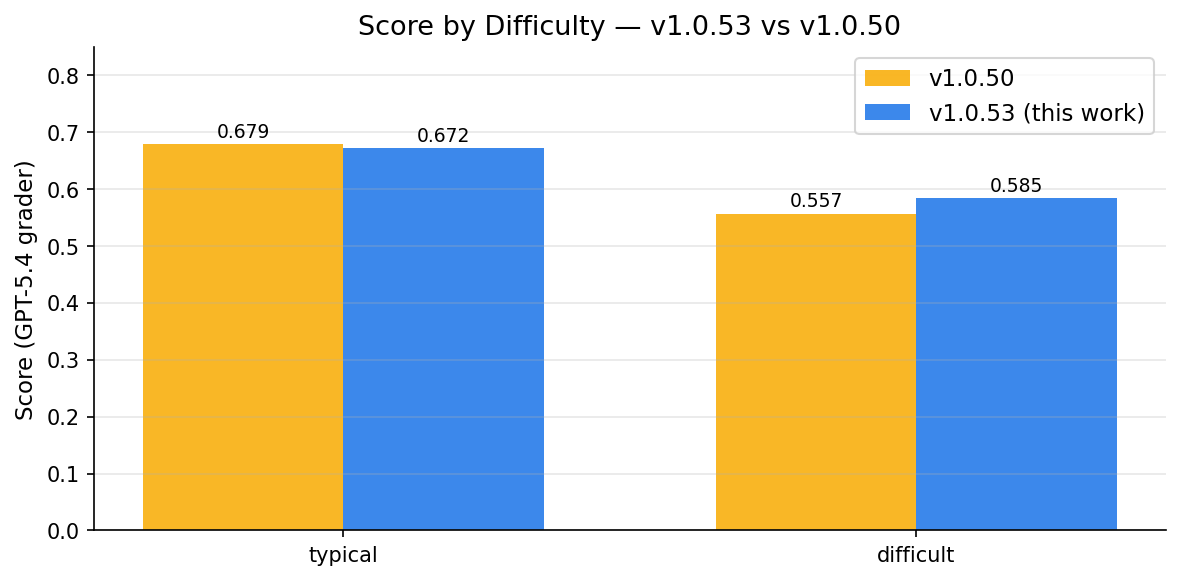}}

}

\caption{\label{fig-difficulty}Score by difficulty category for v1.0.53
vs v1.0.50, GPT-5.4 grader.}

\end{figure}

All changes are within bootstrap σ (≈ 0.025--0.035 per category at these
sample sizes). The consult (+5.7 pp) and difficult (+5.8 pp)
improvements are consistent with the search hygiene changes (v1.0.42)
filtering noise on complex multi-step queries --- these are the
categories most sensitive to intake quality. The slight dips in typical
and research are within noise. The corresponding specialty and
difficulty views are shown in Figure~\ref{fig-specialty} and
Figure~\ref{fig-difficulty}.

The behavior is uneven depending on specialty, Ophtho (0.562) and uro
(0.485) remain the lowest-scoring specialties in
Table~\ref{tbl-v1050-specialty-breakdown} and are the primary targets
for specialty reasoner expansion or RAG augmentation.

\begin{longtable}[]{@{}
  >{\raggedright\arraybackslash}p{(\linewidth - 4\tabcolsep) * \real{0.2000}}
  >{\raggedleft\arraybackslash}p{(\linewidth - 4\tabcolsep) * \real{0.1500}}
  >{\raggedleft\arraybackslash}p{(\linewidth - 4\tabcolsep) * \real{0.1000}}@{}}
\caption{v1.0.50 Pro-graded specialty breakdown, selected
specialties.}\label{tbl-v1050-specialty-breakdown}\tabularnewline
\toprule\noalign{}
\begin{minipage}[b]{\linewidth}\raggedright
Specialty
\end{minipage} & \begin{minipage}[b]{\linewidth}\raggedleft
v1.0.50
\end{minipage} & \begin{minipage}[b]{\linewidth}\raggedleft
n
\end{minipage} \\
\midrule\noalign{}
\endfirsthead
\toprule\noalign{}
\begin{minipage}[b]{\linewidth}\raggedright
Specialty
\end{minipage} & \begin{minipage}[b]{\linewidth}\raggedleft
v1.0.50
\end{minipage} & \begin{minipage}[b]{\linewidth}\raggedleft
n
\end{minipage} \\
\midrule\noalign{}
\endhead
\bottomrule\noalign{}
\endlastfoot
rheum & 1.000 & 8 \\
allergy & 1.000 & 2 \\
primary & 0.837 & 7 \\
derm & 0.833 & 25 \\
endo & 0.835 & 20 \\
nephro & 0.829 & 22 \\
genetics & 0.795 & 4 \\
neuro & 0.680 & 43 \\
cards & 0.627 & 43 \\
obgyn & 0.645 & 40 \\
peds & 0.596 & 28 \\
ophtho & 0.562 & 18 \\
uro & 0.485 & 13 \\
\end{longtable}

\subsection{Length-guidance experiment:
v1.0.53}\label{length-guidance-experiment-v1.0.53}

The version v1.0.53 is aimed at guiding the agent the proper response
length. Its key finding is about \textbf{response length}: v1.0.50
responses averaged \textbf{4383 chars}; v1.0.53 responses average
\textbf{2789 chars} --- a reduction of \textasciitilde1600 chars. The
3000-char cap in the synthesizer and verifier bound hard, cutting the
per-sample length penalty by approximately 0.047
(\(2.94 \times 10^{-5} \times 1594\) chars). The GPT-5.4 score
improvement (+1.06 pp) is attributable primarily to this length-penalty
reduction as we see in Table~\ref{tbl-v1053-overall}.

\begin{longtable}[]{@{}
  >{\raggedright\arraybackslash}p{(\linewidth - 6\tabcolsep) * \real{0.3000}}
  >{\raggedleft\arraybackslash}p{(\linewidth - 6\tabcolsep) * \real{0.0800}}
  >{\raggedleft\arraybackslash}p{(\linewidth - 6\tabcolsep) * \real{0.1500}}
  >{\raggedright\arraybackslash}p{(\linewidth - 6\tabcolsep) * \real{0.4000}}@{}}
\caption{v1.0.53 length-guidance experiment
scores.}\label{tbl-v1053-overall}\tabularnewline
\toprule\noalign{}
\begin{minipage}[b]{\linewidth}\raggedright
Grader
\end{minipage} & \begin{minipage}[b]{\linewidth}\raggedleft
Score
\end{minipage} & \begin{minipage}[b]{\linewidth}\raggedleft
Avg len
\end{minipage} & \begin{minipage}[b]{\linewidth}\raggedright
Notes
\end{minipage} \\
\midrule\noalign{}
\endfirsthead
\toprule\noalign{}
\begin{minipage}[b]{\linewidth}\raggedright
Grader
\end{minipage} & \begin{minipage}[b]{\linewidth}\raggedleft
Score
\end{minipage} & \begin{minipage}[b]{\linewidth}\raggedleft
Avg len
\end{minipage} & \begin{minipage}[b]{\linewidth}\raggedright
Notes
\end{minipage} \\
\midrule\noalign{}
\endhead
\bottomrule\noalign{}
\endlastfoot
Gemini 2.5 Pro & 0.6585 & 2789 chars & 525 samples; −1.86 pp vs v1.0.50
(0.6771) \\
GPT-5.4-2026-03-05 & 0.6272 & 2789 chars & 525 samples; +1.06 pp vs
v1.0.50 (0.6166) \newline \textbf{new best} \\
\end{longtable}

\begin{figure}[H]

\centering{

\pandocbounded{\includegraphics[keepaspectratio]{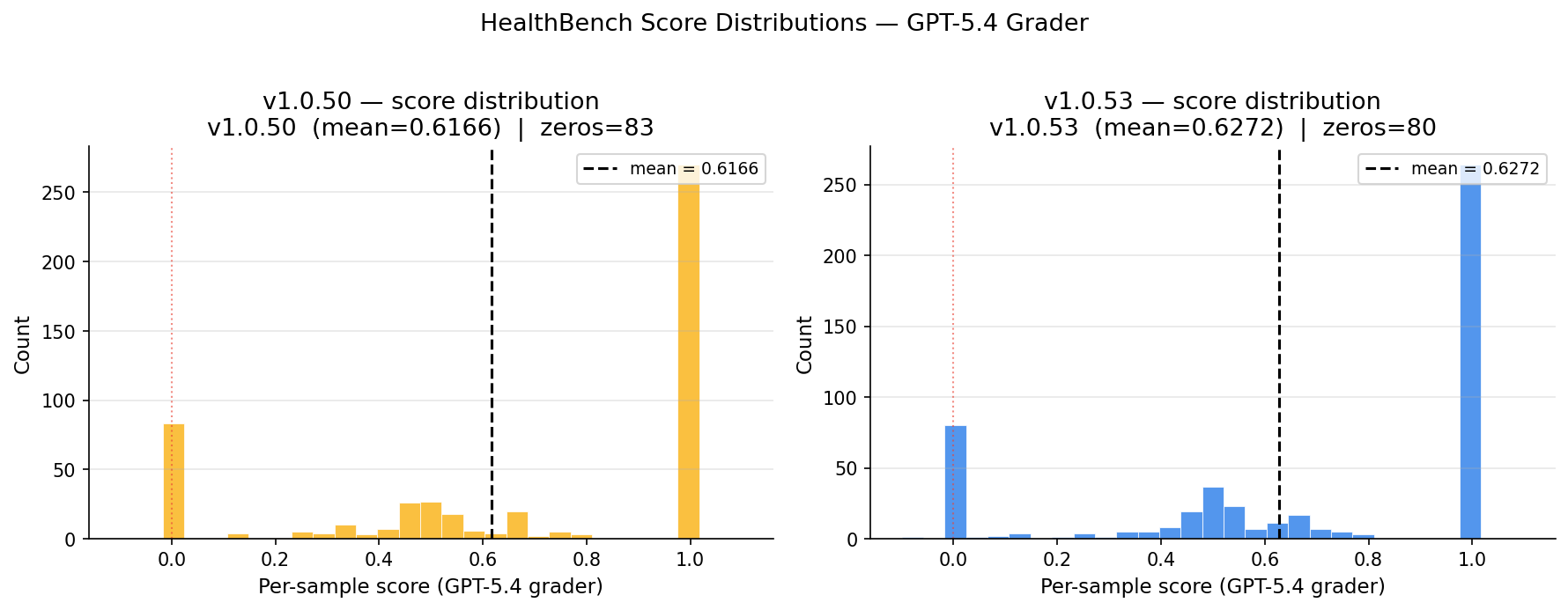}}

}

\caption{\label{fig-distributions}Per-sample score distributions for
v1.0.50 (left) and v1.0.53 (right) under GPT-5.4 grader. v1.0.53 shifts
the distribution rightward; the zero-score spike (length-penalised
failures) is slightly reduced.}

\end{figure}

\begin{figure}[H]

\centering{

\pandocbounded{\includegraphics[keepaspectratio]{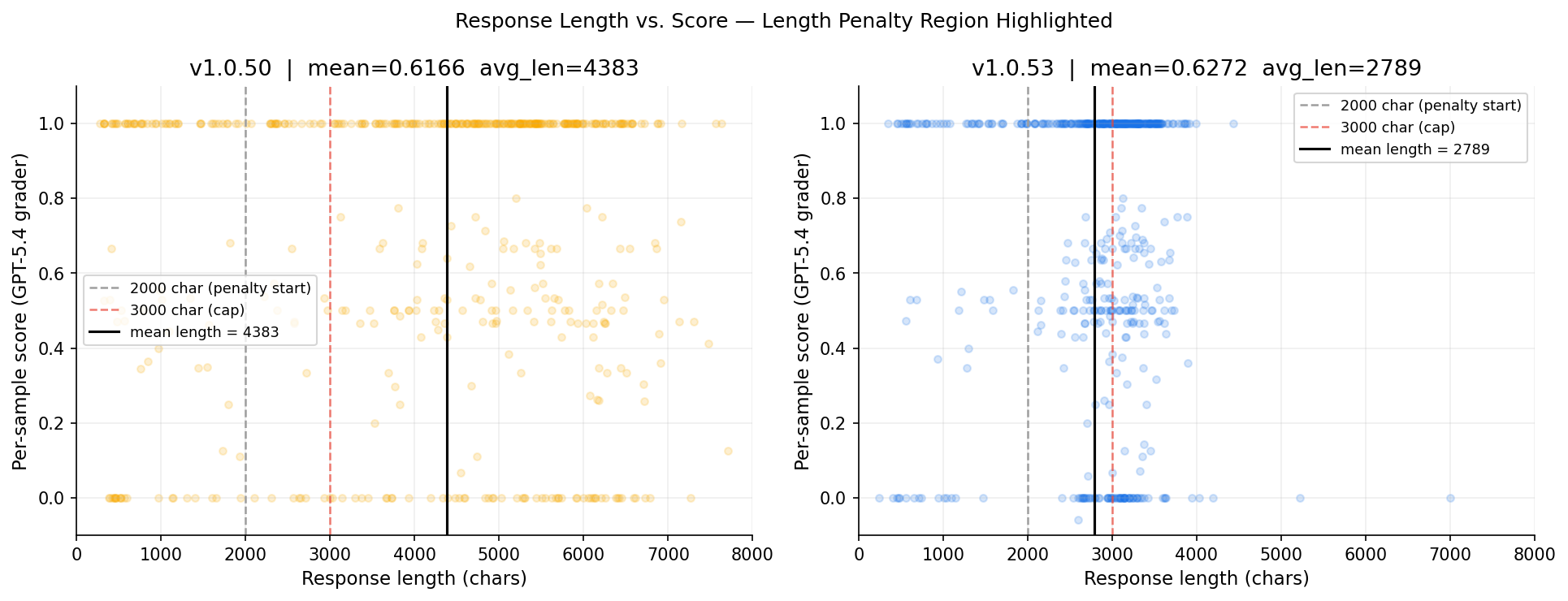}}

}

\caption{\label{fig-length-score}Response length vs per-sample score for
v1.0.50 (left, mean 4383 chars) and v1.0.53 (right, mean 2789 chars).
The 3000-char cap concentrates responses below the steepest
length-penalty region.}

\end{figure}

The Gemini Pro score regressed slightly (−1.86 pp). One interpretation
is that Gemini Pro's rubric is more tolerant of longer, elaborated
responses; cutting length removed content it was rewarding. The two
graders disagree on the direction of the length-content trade-off,
highlighting the inter-grader variance documented in
Section~\ref{sec-multiturn-finding}. The distributional and length-score
diagnostics are shown in Figure~\ref{fig-distributions} and
Figure~\ref{fig-length-score}.

The results with v1.0.53 indicate that the 3000-character length cap is
a net win under the same-instrument comparison (GPT-5.4 +1.06 pp, new
headline best at 0.6272). The cap did not cost rubric anchors ---
zero-scoring cases dropped from 83 (v1.0.50) to 80 (v1.0.53), confirming
the ``Cut verbosity, never content'' rule preserved clinical specifics.
The remaining 80 zeros are content gaps (missing anchor knowledge), not
formatting issues; the right next lever is RAG augmentation
(Section~\ref{sec-guideline-rag}).

\subsection{Benchmark neutrality, fairness concerns and length
correction}\label{benchmark-neutrality-fairness-concerns-and-length-correction}

The comparison with ChatGPT for Clinicians should also be interpreted in
light of benchmark neutrality. OpenAI's HealthBench Professional paper
reports aggregate scores for its own systems and competing models, but
does not publish confidence intervals, per-sample responses, grader
transcripts, or the full set of inference parameters used for each
external model \citep{healthbenchpro2026}. This limits independent
assessment of whether the same evaluation conditions were applied across
systems. The concern is not that the benchmark is intrinsically invalid;
rather, because the benchmark issuer is also a model vendor,
aggregate-only disclosure leaves residual uncertainty about possible
methodology choices that could favor one model family over another
\citep{reuel2024betterbench, sainz2023contamination}. Publishing
per-sample outputs, model settings, prompt templates, flattening
strategy, grader configuration, and uncertainty intervals would make the
comparison more neutral and reduce suspicion of skewed metrics.

Our results reinforce this concern: MDIA receives different absolute
scores under GPT-5.4 and Gemini 2.5 Pro grading, and the length-guidance
intervention even shows different directional effects. This raises a
fairness issue for LLM-as-judge evaluation: which judge is fairest, and
how should fairness be measured? Prior work documents self-preference
effects, where judges may favor outputs from their own model family
\citep{panickssery2024selfpreference, wang2023unfair}. Different
graders---and different reasoning settings---may weight clinical detail,
concision, evidence, refusal behavior, and formatting differently. For
clinical benchmarks, single-grader evaluation is therefore insufficient;
robust reporting should include multi-grader sensitivity analysis,
grader-reasoning settings, and, ideally, a clinician-adjudicated subset
to estimate alignment with expert judgment.

The length correction deserves similar caution. Prior work documents
length bias in automatic evaluation and motivates length-controlled
scoring
\citep{hu2024explaining, dubois2024length, chiu2025morebench, saito2023verbosity},
but the clinical setting makes the trade-off less straightforward.
Penalizing every character beyond a 2000-character break-even point may
discourage unnecessary verbosity, yet it may also penalize clinically
appropriate content such as references, caveats, patient-specific
contraindications, or supplementary rationale. In our data, valid
clinical answers often occupy the 2000-3000 character range, and v1.0.50
averaged 4383 characters before length guidance; the fact that GPT-5.4
rewarded shortening while Gemini regressed suggests that the penalty is
not a model-independent proxy for quality. A more clinically grounded
correction would validate the length-quality curve with clinician
reviewers, test whether the effect is asymptotic rather than linear, and
exclude references, appendices, or supplementary material from the
length penalty when those elements improve auditability without changing
the substantive answer.

\section{Engineering process and lessons
learned}\label{engineering-process-and-lessons-learned}

This section is for fellow practitioners. It documents what \emph{did
not} work, because every ``great result'' paper hides the
trial-and-error behind it.

\subsection{Failed approaches}\label{failed-approaches}

For transparency, Table~\ref{tbl-failed-approaches} summarizes the main
negative or non-improving interventions tested during development,
together with the observed failure modes.

\begingroup
\small
\renewcommand{\arraystretch}{1.5}

\begin{longtable}[]{@{}
  >{\raggedright\arraybackslash}p{(\linewidth - 6\tabcolsep) * \real{0.2000}}
  >{\raggedright\arraybackslash}p{(\linewidth - 6\tabcolsep) * \real{0.3000}}
  >{\raggedleft\arraybackslash}p{(\linewidth - 6\tabcolsep) * \real{0.1500}}
  >{\raggedright\arraybackslash}p{(\linewidth - 6\tabcolsep) * \real{0.3500}}@{}}
\caption{Failed approaches and observed failure
modes.}\label{tbl-failed-approaches}\tabularnewline
\toprule\noalign{}
\begin{minipage}[b]{\linewidth}\raggedright
Approach
\end{minipage} & \begin{minipage}[b]{\linewidth}\raggedright
What we tried
\end{minipage} & \begin{minipage}[b]{\linewidth}\raggedleft
Result
\end{minipage} & \begin{minipage}[b]{\linewidth}\raggedright
Why it failed
\end{minipage} \\
\midrule\noalign{}
\endfirsthead
\toprule\noalign{}
\begin{minipage}[b]{\linewidth}\raggedright
Approach
\end{minipage} & \begin{minipage}[b]{\linewidth}\raggedright
What we tried
\end{minipage} & \begin{minipage}[b]{\linewidth}\raggedleft
Result
\end{minipage} & \begin{minipage}[b]{\linewidth}\raggedright
Why it failed
\end{minipage} \\
\midrule\noalign{}
\endhead
\bottomrule\noalign{}
\endlastfoot
Upgrade verifier Flash → Pro & Pin verifier to Gemini 2.5 Pro & 0.4905
(−3.4 pp) & Pro hedges where the rubric demands an explicit
recommendation. \\
5-node serial safety reviewer & Insert a dedicated
\texttt{safety\_reviewer} after \texttt{intake} & regressed at n=50 &
Inserted node's output replaced the orchestrator's dossier in input
payload. \\
Procedural specialty reasoner with replaced output format & Replace
generic 6-section structure with a tighter procedural format & regressed
by −18 pp on red\_teaming & New format suppressed the
adversarial-framing refusal reflex. \\
Drug-safety tool fired only at orchestrator STEP-0 & Force
\texttt{drug\_state\_safety\_check} as the first tool call & regressed &
Displaced the orchestrator's evidence-gathering budget. \\
Broaden safety gate to fire on every task type & Remove the writing-task
scope condition & −13.8 pp (n=50) & Over-refusal on educational /
counter-misinformation tasks. \\
\textbf{Minimal-input two-section rule (v1.0.38)} & Conditional
bracketed-template + decision-support sidebar for thin prompts &
\textbf{−2.5 pp Pro / −2.7 pp GPT-5.4 at n=525} & Trigger condition too
coarse. See Section~\ref{sec-minimal-input}. \\
\textbf{Reasoner upgrade alone, single-turn flatten (v1.0.39)} & Repoint
reasoners at \texttt{gemini-3.1-pro} with \texttt{last\_user} flatten &
\textbf{−1.65 pp Pro / −1.57 pp GPT-5.4} & Existing prompts tuned
against 2.5 Pro's response style. \textbf{Reversed in v1.0.40 once
multi-turn flatten was applied} --- see
Section~\ref{sec-reasoner-swap-alone}. Reasoner swap × flatten not
separable. \\
Eval via single-agent endpoint & Route all evals to a single agent
rather than the multi-agent graph & Correct results for single-node
capability; misses full 7-node behavior & Single-agent endpoint
exercises only the intake orchestrator. Graph routing, safety gating,
synthesis, and verifier do not fire. Fixed in v1.0.50. \\
\end{longtable}

\endgroup

\subsection{What worked, ranked by
EV-per-effort}\label{what-worked-ranked-by-ev-per-effort}

In contrast to the unsuccessful or non-improving interventions, several
changes produced measurable benefits or improved system robustness. To
make the development lessons more actionable, we rank them below by
expected value per unit of implementation effort, distinguishing
low-cost harness or engine fixes from higher-effort architectural
changes.

\begin{enumerate}
\def\labelenumi{\arabic{enumi}.}
\tightlist
\item
  \textbf{Multi-turn flatten correction (v1.0.40)} --- single biggest
  win this revision. Cost: 30 lines of Python + a default-flag flip.
  Lift: \textbf{+6.6 pp Pro / +6.3 pp GPT-5} at n=525. Surfaces existing
  agent capability instead of adding new behavior.
\item
  \textbf{Engine-level retry-on-empty + JSON-fence stripping} --- single
  biggest reliability win. Cost: 50 lines of Go. Lift:
  \textasciitilde+3-4 pp on the headline floor.
\item
  \textbf{Per-model location override + reasoner swap to Gemini 3.1 Pro}
  --- net positive only when paired with multi-turn flatten. Cost: 20
  lines of Go in the resolver. Lift: roughly cancels v1.0.39's
  regression and adds the multi-turn slice headroom.
\item
  \textbf{Drug-state safety gate with educational exemption (v1.0.27)}
  --- concrete, narrow rule. Cost: \textasciitilde150 lines of prompt.
  Lift: +2.2 pp at n=525.
\item
  \textbf{Specialty router (gi / ophtho / neuro)} --- engine-supported,
  additive. Cost: \textasciitilde600 lines of prompt. Lift: built on top
  of the gate for v1.0.36's overall +3.1 pp.
\item
  \textbf{Site-filtered search hygiene (v1.0.42)} --- prompt-side
  enforcement of existing tool capability. Cost: \textasciitilde20 lines
  of intake prompt. Score-neutral on average; qualitative improvement on
  complex queries (consult/difficult +5-6 pp, within noise but
  directionally consistent).
\end{enumerate}

\subsection{Lessons for other teams}\label{lessons-for-other-teams}

The development process produced several practical lessons that may be
useful for other research and engineering teams evaluating multi-agent
clinical systems, particularly when benchmark results depend on harness
configuration, engine reliability, routing behavior, and full-pipeline
execution.

\begin{itemize}
\tightlist
\item
  \textbf{Always disclose the flatten strategy} :It moves the headline
  by \textasciitilde6 pp on identical agent responses. Flatten strategy
  is now logged in every run manifest.
\item
  \textbf{Multi-turn examples are not edge cases}: 22 \% of HealthBench
  Pro is multi-turn; naive flattening silently understates any
  context-aware agent.
\item
  \textbf{Engine reliability dominates per-iteration variance}: A 3.8 \%
  empty-rate eats more headline than most prompt edits add. Fix the
  floor first.
\item
  \textbf{Reasoner swaps are not separable from prompt or flatten
  changes}: v1.0.39 looked like a regression in isolation; the same swap
  in v1.0.40 (with multi-turn flatten) is a net win. Test combined
  changes, not just deltas.
\item
  \textbf{Specialty-specific anchor knowledge beats single-prompt
  generalism}:, but only for the worst-performing specialties. Adding a
  fourth or fifth branch hits diminishing returns fast.
\item
  \textbf{Strategy mismatches are real, but in-prompt heuristics for
  them are dangerous}: v1.0.38's minimal-input rule looked promising at
  n=50 and regressed at n=525 because its trigger condition
  over-applied. The right answer is a separate task-classifier branch,
  not a heuristic embedded in the synthesizer.
\item
  \textbf{n\textless525 previews are decision-misleading}: v1.0.38
  looked like a +3.4 pp win at n=50 and was a −2.5 pp loss at n=525.
  v1.0.39 trended +3.1 pp at n=128 and converged to −1.65 pp at n=525.
  We now require n=525 validation before promoting any prompt or model
  change.
\item
  \textbf{Validate the full pipeline, not just the entry point}: When a
  clinical system is composed of multiple coordinated stages, an
  evaluation that exercises only the initial agent or controller may
  miss downstream behavior. The evaluation harness should verify that
  all intended stages execute, including routing, specialty reasoning,
  safety checks, synthesis, and final verification. In our case, the
  earlier entry-point evaluation (v1.0.41) was consistent with the
  complete workflow evaluation (v1.0.50), but this cannot be assumed:
  downstream stages may change both failure modes and aggregate
  performance at scale.
\end{itemize}

\section{Future work}\label{sec-future-work}

The results above identify several clear opportunities for improving
MDIA beyond the current v1.0.53 configuration. These future directions
fall into four areas: grounding the agent in curated clinical
guidelines, testing alternative frontier reasoners, making prompt
iteration more systematic, and enabling independent cross-system
regrading. Together, these extensions aim to separate architecture-level
gains from grader effects, reduce residual knowledge gaps, and move the
evaluation closer to robust external validation.

\subsection{Guideline retrieval RAG}\label{sec-guideline-rag}

The failure-mode analysis suggests that the most direct next improvement
is guideline-grounded retrieval. In total, \textbf{62.8\% of
zero-scoring cases correspond to true knowledge gaps}: rubric anchors
that the parametric model did not reliably contain or retrieve, such as
Surgical Apgar scoring, post-CCRT dental osteoradionecrosis risk, the
AHA 2017 infective endocarditis prophylaxis update, or the BMJ 2004
Marik and Zaloga enteral-nutrition meta-analysis.

Retrieval-augmented generation \citep{lewis2020rag} is a natural
response to this failure mode. Prior work has shown substantial gains in
clinical guideline adherence: a NICE-guideline RAG system achieved
99.5\% faithfulness and reduced unsafe responses by 67\%
\citep{nicerag2025}, while broader reviews identify RAG as a standard
approach for grounding LLMs in clinical evidence \citep{ragmedical2025}.

This is also a low-friction extension for the current system. The Hydra
Platform already includes \texttt{tietai-knowledge-service}, with hybrid
dense and BM25 search, RRF fusion, cross-encoder re-ranking, 11
healthcare web-source adapters \citep{xiong2024medrag}, and a working
\texttt{ragquery} Hydra LLM-tool wrapper. Connecting the orchestrator to
\texttt{ragquery} over a curated guideline knowledge base is therefore
expected to require \textbf{2--3 days of work}, rather than a major
architectural redesign. Based on the proportion of failures that appear
retrieval-tractable, the expected lift is \textbf{+3--5 pp}.

\subsection{Alternative reasoner
testing}\label{sec-claude-reasoner-test}

The Gemini 3.1 Pro reasoner produced a net gain in v1.0.40, but only
after the multi-turn context correction was applied. This suggests that
reasoner substitution is not a drop-in optimization: it interacts
strongly with prompt structure, context handling, and specialty routing.

A natural next test is \textbf{Claude 4 Sonnet or Claude 4 Opus}
\citep{anthropic2024claude} as the specialty reasoner. These models may
have a different parametric-knowledge profile and could close gaps in
weaker specialties, particularly ophthalmology and urology, where Gemini
remains below the best-performing domains. However, this should be
treated as a structured re-tuning exercise rather than a simple model
swap. The expected effort is \textbf{2--3 days}, including
reasoner-prompt adaptation and full-benchmark validation.

\subsection{Conflict-resolution harness for prompt
iteration}\label{sec-conflict-resolution-harness}

A repeated pattern during development was that local prompt improvements
often introduced regressions elsewhere. To address this, future work
should include a closed-loop conflict-resolution harness inspired by
Wong et al. \citep{promptdistillation2026}, building on the broader
literature of automatic prompt-optimisation frameworks
\citep{zhou2023ape, khattab2023dspy}.

After each prompt edit, the harness would compare the new run against
the prior version, identify newly failing cases, and use a teacher LLM
to propose minimal additive amendments. These amendments would then be
re-evaluated until the prompt converges without introducing avoidable
regressions. This would convert prompt iteration from manual
trial-and-error into a more systematic regression-management process.

\subsection{Cross-system regrade with OpenAI
outputs}\label{sec-cross-system-regrade}

The most informative external validation would be a cross-system,
cross-grader comparison. In practice, this means grading \textbf{both
MDIA and ChatGPT for Clinicians with both graders}: the GPT-5.4 grader
and the Gemini 2.5 Pro grader.

We have access to both graders for MDIA but do not have the per-sample
outputs from ChatGPT for Clinicians. We therefore cannot construct the
full system-by-grader matrix. We invite OpenAI, or any HealthBench
Professional reporter, to publish per-sample responses so the community
can perform independent regrading and quantify model performance
separately from grader effects.

\section{Conclusion}\label{conclusion}

MDIA achieves stronger HealthBench Professional performance than
OpenAI's flagship medical model, ChatGPT for Clinicians, by combining a
general-purpose LLM with an agentic graph architecture. However,
although HealthBench Professional is designed around realistic clinical
scenarios and explicit benchmark criteria, our results also show that
scores can vary substantially depending on the grader model used. This
reinforces the need to interpret benchmark results as technical
indicators rather than definitive measures of clinical readiness.

This work makes four main contributions:

\begin{enumerate}
\def\labelenumi{\arabic{enumi}.}
\item
  \textbf{A functioning multi-agent clinical pipeline.} MDIA nominally
  exceeds ChatGPT for Clinicians under GPT-5.4 grading on the full
  benchmark (n = 525), with a +3.72 pp margin that remains within
  bootstrap σ and should therefore be interpreted cautiously. More
  substantially, it outperforms the GPT-5.4 single-agent baseline by
  +14.62 pp and physician-written responses by +19.02 pp.~These gains
  are achieved through graph architecture, multi-turn handling, and
  prompt design over an off-the-shelf Gemini 3.1 Pro reasoner, with no
  fine-tuning or privileged retrieval.
\item
  \textbf{A multi-turn context finding.} HealthBench Professional
  performance depends materially on the flattening strategy used in the
  evaluation harness, with an approximately 6 pp effect at n = 525. We
  therefore recommend that future evaluations disclose the flattening
  strategy and default to \texttt{multiturn} for context-aware agents.
\item
  \textbf{An engineering contribution.} Five engine-level reliability
  fixes in the Hydra Platform subagent-graph executor recovered
  approximately 3--4 pp of headline performance previously lost to
  infrastructure instability. In particular, the empty-response rate
  decreased from 3.8\% to 0.2\%.
\item
  \textbf{A validated multi-agent graph architecture.} v1.0.50 was the
  first evaluation run through the correct graph endpoint, confirming
  that specialty routing, drug-state gating, output synthesis, and final
  verification execute sequentially and produce results consistent with
  the prior single-node orchestrator evaluation.
\end{enumerate}

Several areas remain open for future work, including guideline-based RAG
(Section~\ref{sec-guideline-rag}), alternative reasoner testing
(Section~\ref{sec-claude-reasoner-test}), a conflict-resolution harness
(Section~\ref{sec-conflict-resolution-harness}), and a cross-system
regrade if ChatGPT for Clinicians outputs become publicly available
(Section~\ref{sec-cross-system-regrade}).

Finally, the benchmark itself should be treated as an evaluation
instrument whose neutrality depends on disclosure. HealthBench
Professional is valuable because it uses realistic clinician
conversations and explicit rubrics, but aggregate scores alone are not
enough to resolve fairness questions when a model vendor evaluates its
own system against competitors. Future benchmark reports should publish
per-sample outputs, model parameters, grader settings, confidence
intervals, and length-penalty details, and should test whether the
chosen grader and response-length correction remain fair under
independent LLM judges and clinician review.

\section*{Reproducibility}\label{reproducibility}
\addcontentsline{toc}{section}{Reproducibility}

All metrics reported in this paper are reproducible by running the
published graph version on the 525-sample HealthBench Professional
dataset using the documented grader and flattening strategy
(\texttt{multiturn}). We have made the empirical analysis results
publicly available in the TietAI Evals Public repository
\emph{https://github.com/tietai/tietai-evals-public}
\citep{tietaievalspublic2026}.

On the other hand, the following artifacts are available from the
authors upon request:

\begin{itemize}
\tightlist
\item
  \textbf{Graph definitions}: versioned prompt sets for
  v1.0.\{27,36,38,39,40,41,50,53\}, including all node prompts and graph
  topology
\item
  \textbf{Execution Engine}: the Hydra Platform subagent-graph executor,
  including the reliability fixes described in
  Section~\ref{sec-engine-reliability}
\item
  \textbf{Eval harness}: the \texttt{tietai-evals} benchmark runner with
  \texttt{healthbench} and \texttt{regrade} commands, default flatten =
  \texttt{multiturn}
\item
  \textbf{Per-sample grader transcripts}: HTML/JSONL reports for each
  version under both graders (Gemini 2.5 Pro and GPT-5.4-2026-03-05),
  covering all versions from v1.0.27 through v1.0.53 at n=525
\end{itemize}

\section*{Acknowledgements}\label{acknowledgements}
\addcontentsline{toc}{section}{Acknowledgements}

The authors thank the TietAI clinical team for feedback on governance
requirements and the Hydra Platform engineering team for integration
support.

\section*{Conflicts of Interest}\label{conflicts-of-interest}
\addcontentsline{toc}{section}{Conflicts of Interest}

All authors are employees of TietAI, the company that develops and
operates the Hydra Platform and MDIA solution evaluated in this work.
This affiliation may pose a potential conflict of interest in the
design, implementation, interpretation, and reporting of the results. To
support a neutral interpretation, we report both favorable and
unfavorable findings, disclose the evaluation setup and grader
sensitivity, and provide run-level evaluation outputs for independent
inspection. The results should therefore be interpreted as a technical
evaluation of a TietAI-developed system, rather than as independent
clinical validation or definitive evidence of deployment readiness.

\bibliographystyle{unsrt}
\nocite{*}
\bibliography{refs}
\clearpage

\section*{Annex I: Complete model
ranking}\label{annex-i-complete-model-ranking}
\addcontentsline{toc}{section}{Annex I: Complete model ranking}

The comparison of latest MDIA versions under OpenAI's grader
(GPT-5.4-2026-03-05 low) along with the measures disclosed in the
original benchmark paper \citep{healthbenchpro2026} is reproduced in
Table~\ref{tbl-annex-headline-results}.

\begingroup
\small

\begin{longtable}[]{@{}
  >{\raggedright\arraybackslash}p{(\linewidth - 6\tabcolsep) * \real{0.5000}}
  >{\raggedleft\arraybackslash}p{(\linewidth - 6\tabcolsep) * \real{0.2000}}
  >{\raggedleft\arraybackslash}p{(\linewidth - 6\tabcolsep) * \real{0.1000}}
  >{\raggedright\arraybackslash}p{(\linewidth - 6\tabcolsep) * \real{0.2000}}@{}}
\caption{Annex headline same-grader
comparison.}\label{tbl-annex-headline-results}\tabularnewline
\toprule\noalign{}
\begin{minipage}[b]{\linewidth}\raggedright
System
\end{minipage} & \begin{minipage}[b]{\linewidth}\raggedleft
Score
\end{minipage} & \begin{minipage}[b]{\linewidth}\raggedleft
Avg len
\end{minipage} & \begin{minipage}[b]{\linewidth}\raggedright
Reference
\end{minipage} \\
\midrule\noalign{}
\endfirsthead
\toprule\noalign{}
\begin{minipage}[b]{\linewidth}\raggedright
System
\end{minipage} & \begin{minipage}[b]{\linewidth}\raggedleft
Score
\end{minipage} & \begin{minipage}[b]{\linewidth}\raggedleft
Avg len
\end{minipage} & \begin{minipage}[b]{\linewidth}\raggedright
Reference
\end{minipage} \\
\midrule\noalign{}
\endhead
\bottomrule\noalign{}
\endlastfoot
\textbf{MDIA v1.0.53} (Hydra Platform, length-guided synthesizer +
verifier) & \textbf{0.6272} & 2789 & this work \\
MDIA v1.0.50 (Hydra Platform, Gemini 3.1 Pro , graph endpoint) & 0.6166
± 0.0230 & 4383 & this work \\
MDIA v1.0.41 (Hydra Platform, single-agent endpoint) & 0.5775 ± 0.0235 &
--- & this work \\
ChatGPT for Clinicians (best in OpenAI paper) & 0.590 & --- & OpenAI
2026 \\
GPT-5.4 base & 0.481 & --- & OpenAI 2026 \\
Claude Opus 4.7 & 0.470 & --- & OpenAI 2026 \\
GPT-5 & 0.462 & --- & OpenAI 2026 \\
GPT-5.2 & 0.459 & --- & OpenAI 2026 \\
Gemini 3.1 Pro & 0.438 & --- & OpenAI 2026 \\
Physician-written baseline & 0.437 & --- & OpenAI 2026 \\
Grok 4.20 & 0.361 & --- & OpenAI 2026 \\
\end{longtable}

\endgroup

Under OpenAI's own grader on the same 525 HealthBench Professional
cases, MDIA v1.0.53 is the highest-scoring system in this comparison
table. The margin is:

\begin{itemize}
\tightlist
\item
  \textbf{+26.62 pp over Grok 4.20} (0.6272 vs 0.361)
\item
  \textbf{+19.02 pp over the physician-written baseline} (0.6272 vs
  0.437)
\item
  \textbf{+18.92 pp over Gemini 3.1 Pro} (0.6272 vs 0.438)
\item
  \textbf{+16.82 pp over GPT-5.2} (0.6272 vs 0.459)
\item
  \textbf{+16.52 pp over GPT-5} (0.6272 vs 0.462)
\item
  \textbf{+15.72 pp over Claude Opus 4.7} (0.6272 vs 0.470)
\item
  \textbf{+14.62 pp over the GPT-5.4 single-agent system} (0.6272 vs
  0.481)
\item
  \textbf{Nominally +3.72 pp ahead of ChatGPT for Clinicians} (0.6272 vs
  0.590)
\end{itemize}

The ChatGPT for Clinicians comparison remains the most important and the
least certain: the 3.72 pp margin is within bootstrap σ, and OpenAI does
not disclose per-sample outputs, confidence intervals, flattening
strategy, or full inference settings for that system. For this reason,
the result should be read as a same-grader directional comparison rather
than definitive evidence of superiority; see the grader-caveat
discussion in Section~\ref{sec-multiturn-finding}.

\end{document}